\documentclass[journal]{IEEEtran}
\usepackage{soul}
\usepackage{color}
\usepackage{amssymb}
\usepackage{graphicx}
\usepackage{subfigure}
\usepackage{threeparttable}
\usepackage{url}
\usepackage{multirow}

\usepackage{algorithm}    %algorithm package
\usepackage{algorithmic}   %algorithm package
\usepackage{multirow}       %algorithm package
\usepackage{amsmath}      %algorithm package
\usepackage{xcolor}     %algorithm package

\usepackage{booktabs}    %table package
\usepackage{array}
\newcolumntype{I}{!{\vrule width 1pt}}  %set table line width
\usepackage{mdwlist}  %%%%%set description

\begin{document}
\title{Tri-Subject Kinship Verification:\\ Understanding the Core of A Family}

\author{Xiaoqian~Qin,
        Xiaoyang~Tan,
        and Songcan~Chen~{~}% <-this % stops a space
\thanks{Copyright (c) 2013 IEEE}
\thanks{Xiaoqian~Qin, Xiaoyang~Tan and Songcan~Chen is with the Department of Computer Science and Technology, Nanjing University of Aeronautics and Astronautics, Nanjing， China.  Corresponding author: x.tan@nuaa.edu.cn.}% <-this % stops a space
\thanks{Xiaoqian~Qin is with Huaiyin Normal University.}% <-this % stops a space
\thanks{Manuscript received xxx xx, 2015; revised xxx xx, 2015.}}


\markboth{Submitted to IEEE Trans. M.M.}%
{Shell \MakeLowercase{\textit{et al.}}: Bare Demo of IEEEtran.cls for Journals}

\maketitle

\begin{abstract}
One major challenge in computer vision is to go beyond the modeling of individual objects and to investigate the bi- (one-versus-one) or tri- (one-versus-two) relationship among multiple visual entities, answering such questions as whether a child in a photo belongs to given parents. The child-parents relationship plays a core role in a family and understanding such kin relationship would have fundamental impact on the behavior of an artificial intelligent agent working in the human world. In this work, we tackle the problem of one-versus-two (tri-subject) kinship verification and our contributions are three folds: 1) a novel relative symmetric bilinear model (RSBM) introduced to model the similarity between the child and the parents, by incorporating the prior knowledge that a child may resemble a particular parent more than the other; 2) a spatially voted method for feature selection, which jointly selects the most discriminative features for the child-parents pair, while taking local spatial information into account; 3) a large scale tri-subject kinship database characterized by over 1,000 child-parents families. Extensive experiments on KinFaceW, Family101 and our newly released kinship database show that the proposed method outperforms several previous state of the art methods, while could also be used to significantly boost the performance of one-versus-one kinship verification when the information about both parents are available.
\end{abstract}


%in which each feature in an image first estimates its discriminative capability for kinship verification, and uses this to cast a vote for the local region they belong to, finally the local regions are ranked and selected according to the number of votes they receive

% \fi
\IEEEpeerreviewmaketitle

\section{Introduction}
%\IEEEPARstart{T}{his} demo file is intended to serve as a ``starter file''
%for IEEE journal papers produced under \LaTeX\ using
%IEEEtran.cls version 1.8 and later.
%% You must have at least 2 lines in the paragraph with the drop letter
%% (should never be an issue)
%I wish you the best of success.

%\hfill December 27, 2012

Kinship verification from facial images is an emerging problem in computer vision. From an aspect of face recognition, kinship provides us with a valuable and operational opportunity to construct useful relationship between persons based on their visual signals, thus deepening our understanding on their semantics. Applications of kin relationships include face image retrieval \cite{jiang2013query} \cite{benavent2013multimedia} \cite{spyromitros2014comprehensive}    /annotation\cite{ma2013multimedia} \cite{zoidiperson}/organization,  increasing face recognition rates  \cite{wang2010seeing} \cite{shao2014identity}, social media analysis \cite{zhou2012gabor} \cite{xu2014social}, finding of missing children, children adoptions \cite{guo2012kinship}, and so on.

Besides its wide applications, kinship learning is also motivated by the long-term goal of computer vision to go beyond the understanding of a single visual entity (e.g., ``whose face is this?") and to investigate the bi- or tri- relationship among multiple visual entities, e.g., answering such questions as whether a child in a photo belongs to given parents. Actually, recent research has demonstrated that computer vision algorithms have been able to understand individual face image fairly well - the best result on the challenging LFW (labeled face in the wild) face verification database has reached an accuracy as high as 99.15\% \cite{sun2014deep} - even better than what can be done by a human being. However, extending those techniques to characterise the complex relationship among multiple entities is not trivial. One major reason is due to the fact the appearance gap encountered in a kinship problem is much larger than that in a conventional face recognition setting (e.g, given two face images with different sex and different ages, verify whether those two subjects are father and daughter).

In this sense, kinship learning is a step towards such a trend to capture mutual information among different visual entities, particularly multiple face images. Most of current researches \cite{fang2010towards}\cite{xia2011kinship}\cite{xia2012toward}\cite{LuPAMI14}\cite{DibekliogluICCV2013}\cite{yan2014discriminative}, however, mainly focus on the kinship involving only two subjects (one-versus-one) such as father-son or mother-daughter, while in practice, kin relationship involving more subjects are desirable, for example, in the problem of finding missing children, usually we have the photos of both parents, and there is no reason preventing us from using images of both parents at the same time for more effective kinship verification. As another application scenario of law enforcement, it would be beneficial to match the image of a criminal suspect with those of his/her parents to improve the performance of suspect searching. Motivated by this, \cite{ghahramani2014family} assembled a family database containing 45 families with an average of 120 near frontal facial samples per family. Fang et. al. \cite{fangkinship} collected the Family101 kinship dataset, containing 14,816 face images from 206 nuclear families. Both \cite{ghahramani2014family} and \cite{fangkinship} ask questions concerning more general family membership (one-versus-multi) beyond father and son.

In this paper we focus on the problem of tri-subject (one-versus-two) kinship learning (i.e., son-parents and daughter-parents). This is an important special case of the more ambitious one-versus-multi verification and is largely overlooked in literatures. The child-parents is the core and the most basic unit formed in a family and understanding such kind of kin relationship would have fundamental impact on the behavior of an artificial intelligent agent working in a human world. Furthermore, compared to the problem of one-versus-multi kinship verification, the one-vs-two verification is a more convenient and more practical choice - not only because its scope is more controllable, but also because the problem by itself is easier to define since otherwise it could be difficult to determine kinship relations in a big family genetically and without ambiguity, especially for those people among whom the kinship ties are weak.

To address this problem of tri-subject kinship verification, the key idea of our method is to fully exploit the dependence structure between child and parents in a few aspects: similarity measure, feature selection and classifier design. This is based on the observation that compared to the case with only one image from one of the parents, images from both parents could provide richer information about the kinship relation regards to a child, due to the genetic overlapping between a pair of parents and their child. To this end, our contributions are three folds. First, we use a bilinear function to model the similarity between the parents and the child, with the dependence between them captured by a covariance-like matrix learnt from the data. To make this more robust, we introduce a novel relative bilinear similarity model which effectively incorporates the prior knowledge \cite{alvergne2007differential} that children may resemble a particular parent more than another.

Second, we propose a spatially voted method for feature selection, which jointly selects the most discriminative features for the child-parents pair, while taking local spatial information into account. Compared to traditional group-based feature selection methods such as group lasso, we essentially allow the features in a whole image to compete with each other and then select the group in which higher portion of individual features in the corresponding local region win. By contrast, in group lasso, features are teamed together beforehand and have to compete with others as a group. Our method is more flexible than the latter in the sense that it permits fine-grained control over the contribution of each feature to the establishment of one-vs-two kin relationships.

Finally, we release a new face database specific to the tri-subject kinship problem, characterized by over 1,000 child-parents groups. State-of-the-art results are achieved using our method. Interestingly, our experimental results also show that the accuracy of one-vs-one bi-kinship verification benefits a lot by reformulating it as a specific case of one-vs-two tri-kinship verification when the information about a second parent is available.

%\begin{figure}[!htb]
%\centering
%\includegraphics[width=8cm]{./img/p5}
%\caption{ Illustration of the tri-subjects kinship verification problem, where images from the first four groups are with valid child-parents kin relationship respectively, while those in the last group are not, and our goal is to make such a prediction automatically given two images from parents and one image from a child.}
%\label{fig:the aim of family-kinship}\vspace{-3mm}
%\end{figure}

This journal paper builds on the earlier conference work \cite{qin2015kinship}. In what follows we briefly review some of the related work in Section 2, and detail our proposed method in Section3. Our new kinship database is described in Section 4 and experimental results are given in Section 5. We conclude this paper in Section 6.

\section{Related Work}\label{sec_relatedwork}

The aim of bi-subject (one-versus-one) kinship verification through computer vision is predicting whether a given pair of images has kin relation. The research in the field of human visual signal processing \cite{dal2006kin}\cite{debruine2009kin} has provided strong evidence that facial appearance is a useful cue for genetic similarity, since children look more similar to their parents than other adults of the same gender. To find such distinguishable cues from facial appearance, in an early attempt, Fang et al. \cite{fang2010towards} used various features including the skin, hair and eye color, facial structure measures and local/holistic texture.

Later, researchers evaluated various types of feature descriptors for kinship verification. In \cite{guo2012kinship}, the DAISY descriptors are adopted to facilitate local facial patches matching for eyes, mouth and nose with spatial Gaussian kernels.  In \cite{zhou2011kinship}, a spatial pyramid learning-based feature descriptor is utilized to represent kinship faces. In \cite{dehghanlook}, a gated autoencoder method is used to encode the resemblance between a parent and a child, which is trained through minimizing the reconstruction error given a set of randomly sampled local patches. In \cite{shao2014identity}, dense stereo matching is used to determine kinship similarity. Other feature sets for kinship verification include Gradient Orientation Pyramid (GGOP) \cite{zhou2012gabor}, Self Similarity Representation (SSR) \cite{kohli2012self} and prototype-based discriminative feature learning (PDFL) method \cite{yanprototype}. Since semantic-related feature sets such as attributes usually show more tolerance to appearance changes, they are naturally used for kinship verification \cite{xia2012toward}. Based on the idea that people look more like their parents when they smile, \cite{DibekliogluICCV2013} proposes to describe facial dynamics and spatio-temporal appearance over smile expression and uses these to improve the kinship verification rate.

In \cite{LuPAMI14} the authors show that combining several types of middle-level features is useful. For that purpose, they introduced a multiview neighborhood repulsed metric learning method (MNRML) by learning a distance metric under which the samples with a kinship relation are pulled close and those without a kinship relation are pushed away. \cite{yan2014discriminative} and \cite{hu2014large} extract multiple features to characterize face images and maximize the correlation of different features to exploit complementary information for kinship verification. Another way to reduce the appearance similarity gap is to use intermediate samples which bridge the two sides with large divergence. In \cite{xia2011kinship}, \cite{shao2014generalized}, \cite{xia2012understanding} and \cite{syed2014understanding}, such a bridge is constructed by facial images of parents at the similar ages of their children. However, it is not easy to collect such an image set in practice.

While most of the above works focus on the bi-subject (one-versus-one) kinship verification, \cite{ghahramani2014family} and \cite{fangkinship} deal with the one-versus-multi kinship relation. Particularly Ghahramani et al. \cite{ghahramani2014family} addresses the problem of family verification, i.e., predicting whether a query face image has kin relation with multiple family members, by fusing similarity of each member's facial image segments. Fang et al. \cite{fangkinship} tackle the more general family membership classification, i.e., given a query face image, asking which family it belongs to, and they do this with a minimum sparse reconstruction method. Despite the partial success of these methods, we argue that in general it is difficult to establish the relationship between a subject and some members of his/her family through the face appearance if the kinship ties between them is weak\footnote{For example, it makes no sense to reconstruct a man's face image using his father-in-law's.}. Instead we focus on the verification of the most basic unit that forms a family, that is, the child-parents (one-versus-two) relationship. We call this tri-subject kinship verification. The methods developed here can be potentially extended to handle more complex relationship by treating a family tree as an ensemble of tri-relationships.

\section{Tri-Subject Kinship Verification}\label{sec_approach}

In this section, we present our method for tri-subject kinship verification. Assume that we are given a set of $N$ training samples $\{(x_{fi},x_{mi},x_{ci}, y_i)\}_{i=1}^N$, where $x_{fi},x_{mi},x_{ci}\in R^d$ respectively denotes the $i$-th sample of a father, a mother and a child, $d$ is the dimension of the feature representation of a sample, and $y_i\in \{+1,-1\}$ indicates whether this child has a valid kinship relation with the corresponding two parents. Here by kinship relation we mean a very close family type relation, that is, the child is produced by the two parents.

Our goal is to learn a function $f: (x_f,x_m,x_c)\rightarrow \{+1,-1\}$ from the training data to check whether such a kinship could be established for three previously never seen images $(x_f,x_m, x_c)$ of a couple and a child. For simplicity we assume that the gender of both parents images $(x_f,x_m)$ are known and that they indeed genetically produce some children, but we do not know whether $x_c$ is one of them. We also assume that the gender of the test image $x_c$ of the child is known.

\subsection{Two Bilinear Models for Kinship Verification }\label{subsec_bilinear}

\begin{figure}[!t]
\centering
\includegraphics[width=9cm]{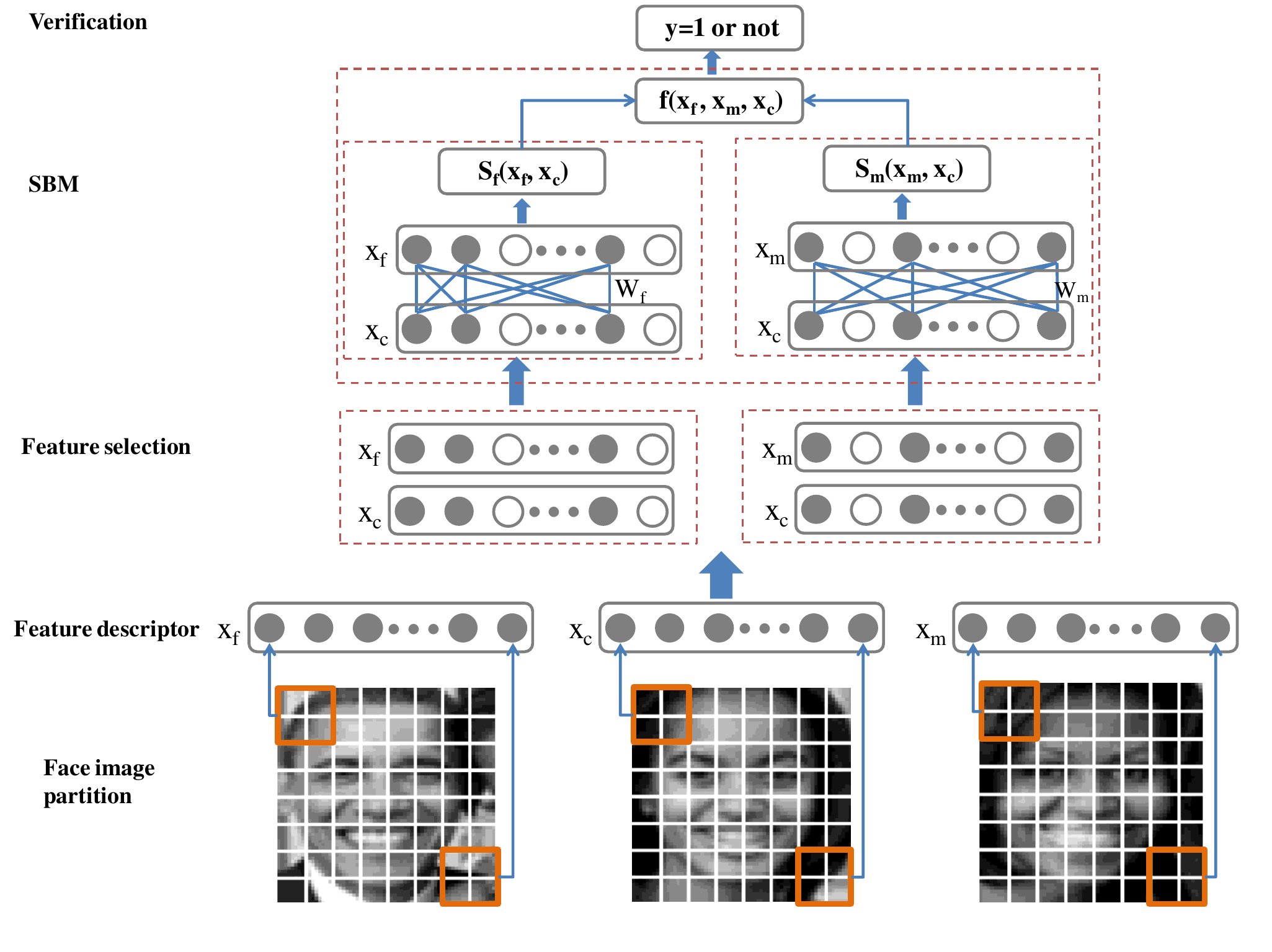}
\caption{The overall architecture of the proposed method.}
\label{fig:the diagram of our method}\vspace{-4mm}
\end{figure}

The overall architecture of the proposed method is shown in Figure~\ref{fig:the diagram of our method}, which can be roughly divided into three stages. Particularly, in the first stage, we partition an image into overlapping patches and extract a middle-level feature descriptor (e.g., $128$-dimensional SIFT features) from each patch (location), which are then concatenated into a feature vector as the input to the next stage. In the second stage we use a spatially voted feature selection method to select the most discriminative local facial patches to improve the robustness. Finally in the third stage, we learn the similarity between parents and child using bilinear models, based on which the final kinship verification is made.

In this work, we explore two ways to encode the similarity between parents and a child. The first one is to decompose the triples of $(x_f,x_m,x_c)$ into two pairs $(x_f,x_c)$ and $(x_m,x_c)$, and the pairwise similarity between them is respectively,
\begin{equation}\label{eq_SBM}
\begin{split}
& s_f(x_f,x_c)=(x_f)^TW_fx_c\equiv \langle x_f,x_c\rangle_{W_f} \\
& s_m(x_m,x_c)=(x_m)^TW_mx_c\equiv \langle x_m,x_c\rangle_{W_m} \\
\end{split}
\end{equation}
where $\langle a,b\rangle_W\equiv a^TWb$, and the transformation matrix $W_f, W_m$ essentially encodes the ``covariance" relationship between a parent and a child, to be learnt from the training data.

Since both $W_f, W_m$ are $d\times d$ matrix and the similarity function is a bilinear function, we call this Symmetric Bilinear Model (SBM). The bilinear model has many advantages compared to the simple Euclidean-based model: 1) it is a natural choice to model the similarity between two subjects; and 2) it is also a much richer model than a traditional linear model -- actually the bilinear model is related to the Mahalanobis distance (especially when the energy of each feature vector is fixed) and hence it can effectively capture the correlation between any two feature variables. However, the bilinear model is different from the Mahalanobis distance in that its parameter matrix $W$ is not necessarily a positive definite matrix, which not only indicates that it could be more flexible than a traditional metric learning-based method, but also means that what a bilinear model learns is not a metric but a classifier. But this is exactly what we need -- a model to predict directly whether a given pair of subjects has some kind of kinship, rather than the metric between them.

We further denote the probability that a child $x_c$ belongs to a pair of parents $(x_f,x_m)$ as $P(y=1|x_f,x_m,x_c)$, and it is linked to our verification function $f(x_f,x_m,x_c)$ through a sigmoid function, i.e.,

\begin{equation}\label{eq_prob1}
  p(y=1|x_f,x_m,x_c)=\sigma (f(x_f,x_m,x_c))
\end{equation}
where sigmoid function $\sigma$ is defined to be $\sigma (x)=\frac{1}{1+e^{-x}}$. The verification function $f(x_f,x_m,x_c)$ is modeled as the linear combination of two pieces of evidence, i.e., the similarity of $x_c$ to $x_m$ and $x_f$, respectively,
\begin{equation}\label{eq_alaternative}
  f(x_f,x_m,x_c)=\beta_1s_f(x_f,x_c)+\beta_2s_m(x_m,x_c)+b
\end{equation}
where the combination coefficients $\beta_1$ and $\beta_2$ are two scalars and $b$ is the similarity threshold term. To learn these parameters, we maximize the conditional likelihood defined by Eq.~\ref{eq_prob1} by plugging Eq.~\ref{eq_SBM} into it, with $L_2$ regularization added. How to learn the pairwise similarity Eq.~\ref{eq_SBM} will be detailed in the next section.

Alternatively, one can treat the parents and the child as samples from two domains. Let us denote the parents domain as $\mathcal{P}$, with data points $(x_{f1},x_{m1}),(x_{f2},x_{m2}),...,(x_{fN},x_{mN})$, and the child domain as $\mathcal{C}$, with data points $x_{c1},x_{c2},...,x_{cN}$. With these notations, one can model the similarity between a child $x_c$ and his/her parents $x_p = (x_f,x_m)$ as,

\begin{equation}\label{eq_bilinar}
  s_p(x_p,x_c)=\langle x_p,x_c\rangle_{W_p}
\end{equation}
where $W_p$ is a $2d\times d$ matrix. This model is called Asymmetric Bilinear Model (ABM) in what follows.

For the ABM model, our verifier is defined as follows,
\begin{equation}\label{eq_prob2}
  p(y=1|x_f,x_m,x_c)=\sigma(s_p(x_p,x_c)+b)
\end{equation}
where $\sigma$ is the sigmoid function. The parameters $\{W_p,b\}$ are learnt using the following regularized logistic regression objective,
\begin{equation}\label{eq_the model}
%\begin{split}
 \min_{W_p,b } \sum_{i=1}^{N} log(1+exp(-y_{i}(\langle x_{pi},x_{ci}\rangle_{W_p}+b)) +\lambda \Arrowvert W_p \Arrowvert_{*}
%\end{split}
\end{equation}
where $b$ is the threshold, and $\Arrowvert W_p \Arrowvert_{*}$ is the trace norm, defined as $\Arrowvert W_p \Arrowvert_{*}= \sum_{i} \sigma_{i} $ (the  $\sigma_{i}'s$  are the singular values of $W_p$). With appropriate parameter $\lambda$, the trace norm shall force a solution with many singular values of $W_p$ being exactly zero. This allows a more compact representation of the data, thus being useful especially when the original feature space is high-dimensional. Equation~\ref{eq_the model} is a nonsmooth convex objective and one can use proximal methods to solved it, where at each step the singular values of the standard gradient update are replaced by their soft-threshold versions. See \cite{ji2009accelerated} for details on an efficient implementation of this.

Comparing the SBM model and the ABM model, the SBM learns two simple models first (i.e., by learning two $d\times d$ parameter matrices $W_f$,$W_m$ separately) and then combines them with coefficients $\beta_1$ and $\beta_2$, while the ABM learns a bigger model at one time (a $2d\times d$ parameter matrix $W_p$). In other words, the SBM essentially combines two sub-modules (one does the father-child kinship verification and the other for the mother-child relation), which not only makes the learning task easier, but also provides further flexibility to calibrate the outputs of the two sub-modules such that the final prediction (father/mother-child) is as accurate as possible. By contrast, the ABM model tries to do this in one big step, which is much harder especially when the size of dataset is relatively small (less than 2K images for training in our case) for a $2d\times d$ matrix (for $d = 400 $, the total number of parameters would be $2\times 400 \times 400 = 160,000$).

%Comparing the SBM model and the ABM model, they have almost the same number of parameters except that two more parameters, i.e., $\beta_1$ and $\beta_2$, are needed in the case of SBM (c.f., Eq.~\ref{eq_alaternative}), but it decomposes a big asymmetric {\color{blue}{$2d \times d$}} matrix $W_p$ into two small symmetric {\color{blue}{$d \times d$} matrices} (i.e., $W_f,W_m$). This makes the SBM model more efficient to learn and have a better chance to generalize to the unknown data.

\subsection{Learning A Relative Pairwise Similarity Measure}
Note that the SBM model introduced in Eq.~\ref{eq_SBM} is a pairwise similarity model without exploiting the dependence structure among parents and child, which can be considered as a limitation. In fact, one can interpret the SBM as a likelihood model, while to better model the similarity between a father and a son for example, one should put it under the context of three subjects - i.e., instead of modeling the marginal pairwise similarity (e.g., $p($father is similar to son$)$), modeling its conditional version (e.g., $p($father is similar to son$|$father,mother,and son$)$). One major advantage of this is to allow us to embed various prior knowledge concerning tri-subject groups into the similarity model. In this work, we are particularly interested in the prior knowledge that children may resemble a particular parent more than another \cite{alvergne2007differential} - ``Jack looks more like his father than his mother" or ``John has similar appearance with her mother".

Let us denote the probability that a child looks more like his/her father or his/her mother as  $p^{fc}$ and $p^{mc}$ respectively, i.e., `$p^{fc}=1$' means that a child looks more like his father than his mother. Taking the parents as references, the child is either more like his/her father or more like his/her mother, so we have $p^{fc}+p^{mc}=1$. We therefore define the two probabilities using the softmax function, based on the pairwise similarity model defined in Eq.~\ref{eq_SBM},
\begin{equation}\label{eq_scale}
\begin{split}
  p^{fc}=\frac{exp(s_f(x_f,x_c))}{exp(s_f(x_f,x_c))+exp(s_m(x_m,x_c))}\\
  p^{mc}=\frac{exp(s_m(x_m,x_c))}{exp(s_f(x_f,x_c))+exp(s_m(x_m,x_c))}
\end{split}
\end{equation}

Incorporating these into the SBM model, we obtain the following relative symmetric bilinear model (RSBM),
\begin{equation}\label{eq_RSBM}
\begin{split}
& s^R_f(x_f,x_c)=p^{fc}\cdot \langle x_f,x_c\rangle_{W_f} \\
& s^R_m(x_m,x_c)=p^{mc}\cdot \langle x_m,x_c\rangle_{W_m} \\
\end{split}
\end{equation}

One remaining problem is how to determine these priors. Eq.~\ref{eq_scale} shows that they depend on the parameters $W_f$ and $W_m$, which suggests a natural iterative procedure - initialize $p^{fc}$ and $p^{mc}$ first, then optimize $W_f$ and $W_m$ in a supervised manner, finally update $p^{fc}$ and $p^{mc}$ again. In this work, we learn $W_f$ and $W_m$ separately using the same trace-norm regularized logistic regression model as that shown in E.q.~\ref{eq_the model}.

However, updating $p^{fc}$ and $p^{mc}$ is somewhat subtle - the range of the sigmoid function of E.q.~\ref{eq_scale} is in $[0,1]$, meaning that when one of $p^{fc}$, $p^{mc}$ reaches $1$ the other one must be nearly $0$. This is risky, since for the one with $0$ probability, the contribution of its corresponding similarity could be cancelled out. To prevent this from happening, we update the new $p^{fc},p^{mc}$ using a stabilizing term, as follows,
\begin{equation}\label{eq_rectify propability}
\begin{split}
p^{fc}_{new}= \alpha p_0^{fc}+(1-\alpha) p^{fc}_{cur}~~~~~~~~~~~~~~\\
p^{mc}_{new}= \alpha p_0^{mc}+(1-\alpha) p^{mc}_{cur} , ~~0<\alpha<1
\end{split}
\end{equation}
where $\alpha\in (0,1)$ is a trade-off parameter, and the stabilizing terms $p^{fc}_{0}$, $p^{mc}_{0}$ are initialized to be 0.5 for each sample, and $p^{fc}_{cur}$, $p^{mc}_{cur}$ are priors calculated according to the $W_f$ or $W_m$ values estimated in the current iteration. In other words, we choose not to trust the currently-estimated similarity prior too much and always regularize it with some fixed stabilizing value. Principally one can optimize the value of $\alpha$ by plugging Eq.~\ref{eq_rectify propability} into the corresponding  regularized logistic regression objective function while treating $W_f$ or $W_m$ as a constant, but in our implementation we set it using a cross validation strategy\footnote{In practice, a small value of $\alpha=0.1$ usually works well.}.

The proposed RSBM algorithm is summarized in Algorithm \ref{alg:scaled SBM}.

\begin{algorithm}[!htb] %算法的开始
\renewcommand{\algorithmicrequire}{\textbf{Input:}}
\renewcommand\algorithmicensure {\textbf{Output:} }
\caption{Solving the Relative Symmetric Bilinear Model (RSBM)}
\label{alg:scaled SBM}
\begin{algorithmic}[1] %这个1 表示每一行都显示数字
\REQUIRE ~~\\ %算法的输入参数：Input
Training images: $S=\{(x_{fi},x_{mi},x_{ci}, y_i)\}_{i=1}^N$;\\
Parameters: regularization term $\lambda$, iteration number T, and trade-off parameter $\alpha$
\ENSURE ~~\\ %算法的输出：Output
Symmetric transformation matrix $W_f$, $W_m$;\\
\STATE \textbf{Initialization: }
\STATE~~~Decompose S into two sets $S_f=\{(x_{fi},x_{ci}, y_i)\}_{i=1}^N$ and $S_m=\{(x_{mi},x_{ci}, y_i)\}_{i=1}^N$;
\STATE~~~~Set $p_0^{fc}=[\frac{1}{2},\frac{1}{2},...,\frac{1}{2}], p_0^{mc}=[\frac{1}{2},\frac{1}{2},...,\frac{1}{2}]$;
\STATE \textbf{For} $L=1,2,...,T$ \textbf{do}
\STATE~~~~Estimate $W^f$ and $W^m$ by solving regularized logistic regression objective (c.f., E.q.~(\ref{eq_the model}));
\STATE~~~~~Update the pairwise similarity with E.q.~(\ref{eq_RSBM});
\STATE~~~~~Estimate $p^{fc}_{cur},p^{mc}_{cur}$ using E.q. (\ref{eq_scale});
\STATE~~~~~Update $p_{new}^{fc},p_{new}^{mc}$ by using E.q. (\ref{eq_rectify propability});
\STATE~~~~~Set $p^{fc}=p_{new}^{fc}$; $p^{mc}=p_{new}^{mc}$;
\STATE \textbf{end for}
\STATE \textbf{Output} Symmetric transformation $W_f$ and $W_m$.
\end{algorithmic}
\end{algorithm}

\subsection{Spatially Voting for Feature Selection}\label{sec_vote}

%Although the trace term in our objective can be considered as a form of sparse learning but it only does this on the vector of singular values, rather than directly on the second layer of the network.
The total number of parameters (i.e., $W_p$, $W_f$ and $W_m$) for our kinship verification model grows quadratically with the dimensions of input features, hence performing feature selection is needed. It can be observed that some important genetic characteristics for a kinship relationship are distributed locally in face images, and it is better to learn them by finding the most discriminative local facial regions (patches) with some supervised information. Furthermore, we want to select those most discriminative patches from both parents and the child images simultaneously such that good generalization can be obtained.

\begin{figure}[!t]
\centering
\includegraphics[width=9cm]{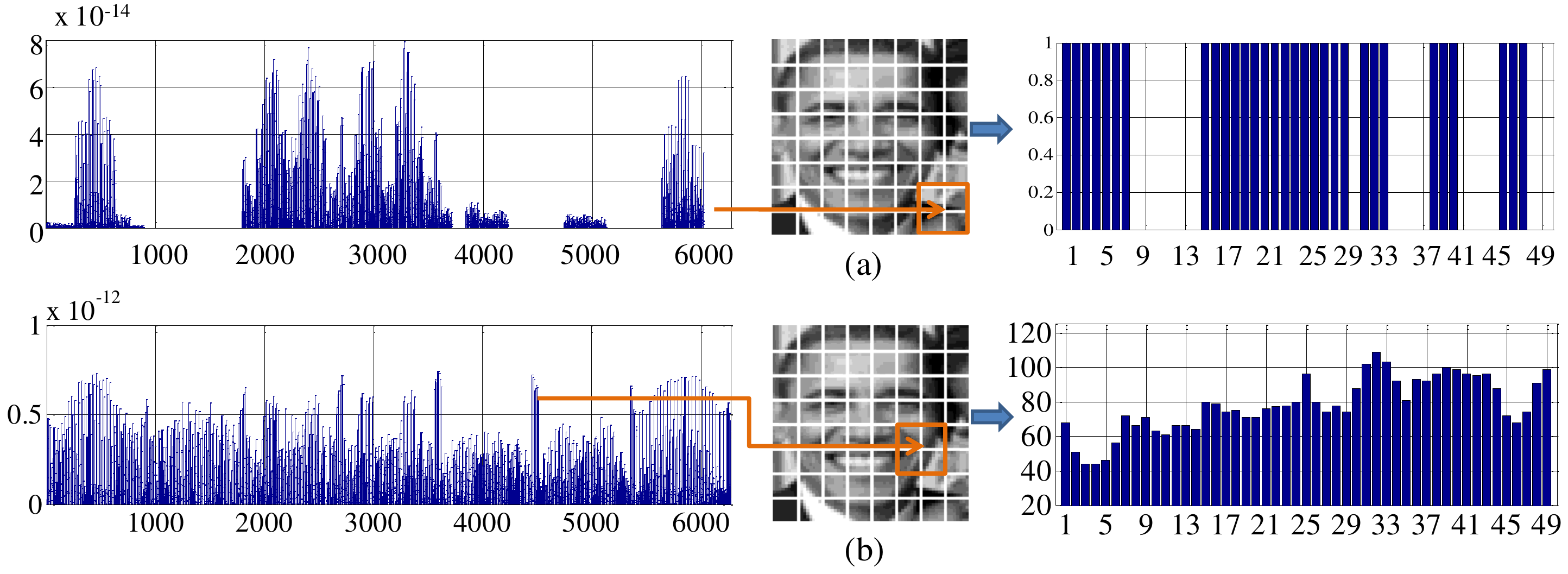}
\caption{Patch selection using (a) the group lasso and (b) the proposed feature selection method. Here for illustration purpose the face image in the middle is partitioned into 49 overlapping patches. For (a), the group lasso method directly selects the most discriminative patches by imposing group competition, and the selected groups (patches) are indicated on the right with blue bars, while the corresponding weights of each feature vector in a group is shown in the left histogram; For (b), the discriminative power of each feature (i.e., the weight vector $u$, see text for details) is first estimated and is shown in the left histogram, while the histogram on the right shows for each patch how many votes it receives, and the first $K$ patches receiving the highest number of votes will be selected.}
\label{fig:patches selection method}%\vspace{-2mm}
\end{figure}

One simple way for this is to treat each patch in an image as a group and use existing techniques such as group lasso \cite{meier2008group} to select a few groups (patches) such that they give the best prediction accuracy, see Fig.\ref{fig:patches selection method} (a) for illustration. However, one drawback of this method is that the feature selection is performed at the level of groups (patches), i.e., the features have to be teamed together before competition and this may hurt the flexibility of feature selection. To overcome this, we adopt an alternative strategy - competition before grouping. That is, all features extracted at each location in a given image (c.f., Section~\ref{subsec_bilinear} on how we extract features) are allowed to freely compete with each other and then select the groups (local regions) in which higher portion of individual features win. Hence our method works in a finer granularity than that of the group lasso. The process of our vote-based feature selection method and group lasso is shown in Fig.\ref{fig:patches selection method}.

Particularly, our algorithm has two steps. In the first step, we evaluate the discriminative power of each feature of a parent regard to the given child. For this we decompose the triple of $(x_f,x_m,x_c)$ into two pairs of $(x_f,x_c)$ and $(x_m,x_c)$. Then for a pair of father-child features $(x_f,x_c)$, we first concatenate them into a $2d$-dimension vector denoted as $a^f$, and learn a weight vector $u^f$ with the same dimension using the following sparse $l_{1}$ regularized logistic regression objective,
\begin{equation}
\begin{split}
 &\min_{u^f} \sum_{i=1}^{N} \log(1+exp(-y_{i}\cdot \langle u^f,a^f_{i}\rangle)+\gamma \Arrowvert u^f \Arrowvert_{1}
\end{split}\label{eq:select dimension}\vspace{-4mm}
\end{equation}
where $y_i=1$ if the pair is a positive sample and $-1$ otherwise. Solving this will give us a $2d$-dimension vector $u^f$ with its first half and the second half respectively representing the importance of each feature of the father and the child. The same procedure is repeated for the mother-child pairs and yields a vector $u^m$.

Now, instead of performing feature selection directly using the information contained in $u$, we use this to vote the patches of face images and select those patches receiving high votes for face representation. Particularly, after solving the L1 logistic regression, we calculate separately for the parent and child how many votes per patch received from $u$. Intuitively, the more votes a patch receives, the more important it is.

Fig.\ref{fig:patches selection method} (b) illustrates this procedure. Since we know the mapping structure between each feature and each patch beforehand, the votes $v_k$ received by the $k$-th patch can be simply calculated as the sum of weights of $u$ corresponding to this patch, i.e., $v_k=\sum_{j\in k}u_j$, where $u_j$ denotes the $j$-th element of vector $u$ corresponding to patch $k$. Note that for a patch of the child image, it would receive votes from the corresponding features of both $u^f$ and $u^m$, while for a father or a mother patch, its vote comes merely from $u^f$ or $u^m$ accordingly. After voting, we select the first $K$ patches with the highest $v_k$ value for parent and child respectively, where $K$ is set using cross validation over a validation set in our implementation (the best value is usually between 20 and 30 with 49 patches per face.).

As mentioned previously, after patch selection, we collect for each image the selected $K$ patches and encode them with SIFT descriptor, which are further concatenated to form a feature vector $x$ for that face.

\section{The TSKinFace Database and Evaluation Protocol} \label{sec_TSK}

\begin{figure}[!t]
\centering
\includegraphics[width=8cm]{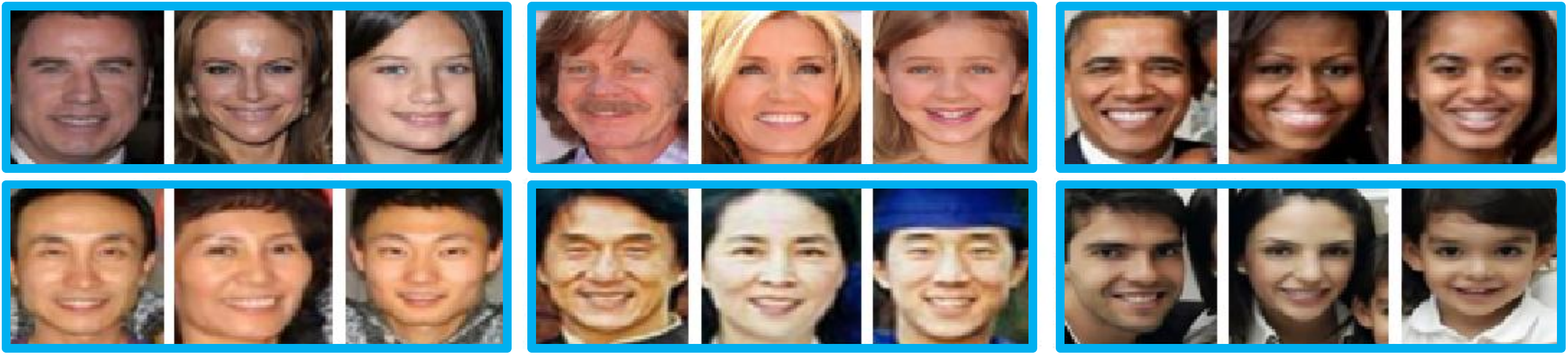}
\caption{Some family image groups of our TSKinFace database, where each group consists of a family triple of a father, a mother and a child. The first row shows three Father-Mother-Daughter (FM-D) relation families, respectively and the second row are three Father-Mother-Son (FM-S) relation families, accordingly.}
\label{fig:TSKinFace pairs image}%\vspace{-3mm}
\end{figure}

To analyze the behavior of the proposed algorithm for tri-subject kinship verification, we constructed a new kinship face database named TSKinFace (Tri-Subject Kinship Face Database). All images in the database are harvested from the internet based on knowledge of public figures family and photo-sharing social network such as flickr.com. During images collecting, we impose no restrictions in terms of pose, lighting, expression, background, race, image quality, etc. Fig.~\ref{fig:TSKinFace pairs image} shows some image groups of child-parents pair from our TSKinFace database. This database will be made publicly available online\footnote{Available at: \emph{http://parnec.nuaa.edu.cn/xtan/data/TSKinFace.html}} to advance the research and applications related to this topic.

Table \ref{tab:difference in database} gives a comparison between our TSKinFace database and other existing kinship databases of human faces. It can be seen that our database is characterized by the largest number of people and families. Specifically, the number of families contained in our database is over 20 times more than that of \cite{ghahramani2014family} and about 5 times more than that of the Family101 database. These features make our dataset particularly suitable for one-vs-two type kinship verification.

\begin{table} [!htb]
\scriptsize
\caption{Comparison of our TSKinFace database and some existing kinship databases of human faces, where ``\#Groups'' refers to the number of kinship relation groups (blood-relation family) in the database, and ``Family structure" refers to the existence of family relationship in the database}
\begin{center}
%\begin{tabular}{p{3.7cm} p{0.5cm}<{\centering} p{0.5cm}<{\centering} p{0.4cm}<{\centering} p{1.9cm}<{\centering}}
\begin{tabular}{ccccc}
\toprule
Database   & \#People     & \#Images	   & \#Groups   &Family structure?\\
\midrule
CornellKin  \cite{fang2010towards} & 300  & 300  & 150   & NO\\
UB KinFace  \cite{shao2011genealogical}\cite{xia2012toward}  & 400  & 600  & 200   & NO\\
KinFaceW-I \cite{LuPAMI14} &1066   &1066   &533   & NO\\
KinFaceW-II \cite{LuPAMI14} & 2000  & 1000  & 1000  & NO\\
Family101 \cite{fangkinship}   &607   &14,816   &206   &YES \\
Database\cite{ghahramani2014family} &--    &5400   & 45   &YES \\
TSKinFace   &2589  &787    & 1015 & YES\\
\bottomrule
\end{tabular}
\end{center}
\label{tab:difference in database}%\vspace{-3mm}
\end{table}

In particular, we are interested in three kinds of child-parents families in real life, i.e., Father-Mother-Daughter (FM-D), Father-Mother-Son (FM-S) and Father-Mother-Son-Daughter (FM-SD). For each type, we collected $274$, $285$ and $228$ family photos respectively, with one photo per family. Using these, we constructed two kinds of family-based kinship relations in the TSKinFace database: Father-Mother-Son(FM-S) and Father-Mother-Daughter(FM-D). The FM-S and the FM-D contain $513$ and $502$ groups of tri-subject kinship relations (c.f., Fig.~\ref{fig:TSKinFace pairs image}), respectively. Hence we have $1015$ tri-subject groups in our database totally. The families included in our database are diverse in terms of races as well. For FM-S relation, there are $343$ and $170$ groups of tri-subject kinship relations for Asian and non-Asian, respectively. And for FM-D relation, the numbers for Asian and non-Asian groups are respectively $331$ and $171$.

%Fig.~\ref{fig:Statistical information of TSKinFace database} gives the statistical information from the perspective of kinship relations and race.
%\begin{figure}[!htb]
%\centering
%\includegraphics[width=8cm]{p2}
%\caption{Statistical information of the TSKinFace database. Left: distribution of tri-subjects kinship relations; Right: distribution of races in terms of Asian vs. Non-Asian.}
%\label{fig:Statistical information of TSKinFace database}\vspace{-3mm}
%\end{figure}

\def\para#1{\medskip\noindent{\bf #1}}
\para{Preprocessing} All downloaded images undergo the same geometric normalization prior to analysis: face detected and cropped using our own implemented Viola-Jones detector, rigid scaling and image rotation to place the centers of the two eyes at fixed positions, using the eye coordinates output from an eye localizer \cite{tan2009enhanced}; image cropping to $64 \times 64$ pixels and conversion to 8 bit gray-scale images. In our experiments, each face image was divided into $7 \times 7$ overlapping patches and the size of each patch is $16 \times 16$. For each patch, we extracted a 128-dimensional SIFT feature. Except mentioned otherwise, for all experiments described in this work, the SIFT is adopted as our default feature descriptor.

\para{Evaluation Protocol} We design a verification protocol for our database following \cite{xia2011kinship} and \cite{LuPAMI14}: the database is equally divided into five folds such that each fold contains nearly the same number of face groups with kinship relation, which facilitates five-fold cross validation experiments. Table \ref{tab:experimental protocol} lists the face number index for the five folds of our TSKinFace database. For face images in each fold, we consider all groups of face images with kinship relation as positive samples, while the negative samples are a random combination with a child image and two parents images subjected to the constraint that the child was not produced by them. In general, the number of negative samples is much more than that of the positive samples. In our experiments, each couple and child images appeared only once in the negative samples. Hence, the number of positive groups and negative groups are the same.
%\end{basedescript}

\begin{table} [!htb]
\caption{Face number index of each fold of the TSKinFace database}
\begin{center}
\begin{tabular}{cccccc}
\toprule
Fold   & 1	& 2	& 3	& 4 &5 \\
\midrule
FM-D   & [1,100]  &[101,200] &[201,300] &[301,400] &[401,502]\\
FM-S   & [1,102] &[103,204]&[205,306] &[307,408] &[409,513]\\
\bottomrule
\end{tabular}
\end{center}
\label{tab:experimental protocol}
\end{table}

\section{Experiments}

\subsection{The Tri-Subject Kinship Verification}
To the best of our knowledge, there are very few works that tackle the tri-subject kinship verification problem, and it is very difficult to find an existing method directly comparable to ours. We therefore design a naive baseline by concatenating the feature vectors of three visual entities and learning a linear SVM for verification. We denote this method as `concatenated+SVM'.

Alternatively, one can use any existing state-of-the-art bi-subject kinship verification model to score the similarity between a child and his/her parents separately, and then train a linear SVM over these to make the final prediction (c.f., E.q.~\ref{eq_alaternative}). Here two best performers (on the KinFaceW dataset) on bi-subject kinship learning, i.e., neighborhood repulsed metric learning (NRML) \cite{LuPAMI14}, and gated autoencoder \cite{dehghanlook} are adopted as our base models. Furthermore, considering that the similarity modeling is related to metric learning, we also include two classical metric learning algorithms, i.e., Information-theoretic
metric learning (ITML) \cite{davis2007information} and  large
margin nearest neighbor classification (LMNN) \cite{weinberger2006distance} as the base models.

Although they deal with different problem, the image set based face verification bears some similarities to the problem of tri-subject kinship verification from the respect of methodology, i.e., both involve similarity matching between multiple faces. Hence in this work, we also adopt one of the best performers on the YouTube Face database, i.e., DDML (Discriminative Deep Metric Learning) \cite{hu2014discriminative}, to score the pairwise similarity between a child and his/her parents, and then train a SVM over these to make the final prediction. Particularly, we train a deep metric learning network with three layers using our own implementation, with the threshold $\tau$, the learning rate $\mu$ and regularization parameter $\lambda$ set to be $3, 10^{-3}, 10^{-2}$, respectively.

Our method is also closely related to Fang et al. \cite{fangkinship} in that both deal with the family structure. However, since their method is mainly designed for kinship classification, it is not directly comparable to ours. But we follow their ideas to build a linear SVM-based kinship verifier to make the comparison feasible. Particularly, we construct a reconstruction errors-based representation (at the patch level) for each face using sparse group lasso \cite{fangkinship}, by treating images belonging to the same family as a group.

Finally, we compare several variants of the proposed method in our experiments, as follows,
\begin{itemize}
  \item With/without feature selection (FS): to investigate the effectiveness of the proposed vote-based feature selection method, for both SBM and ABM models, we evaluate their with/without feature selection versions.
%  \item Working at the block level: The proposed method can also be applied at the level of blocks (patches), i.e., making series of verification predictions first based on the patches selected, and then aggregate these meta-decisions for the final judgement.
  \item Working at the block level: The proposed method can also be applied at the level of blocks (patches), i.e., selecting the most discriminative patches first, then learning the ``covariance" relationship and making verification predictions based on each selected patches, and finally aggregating these meta-decisions through linear SVM for the final verification judgement.
\end{itemize}
In what follows, a notation like `RSBM-block-FS' means a Relative Symmetric Bilinear Model (RSBM) with spatially voted feature selection (FS), working at the block level.

Unless otherwise noted, in all experiments we use the following default parameter settings: $\lambda = 5.0$ in Eq.~\ref{eq_the model} (but change to 0.1 if working at the block level); $\gamma = 0.08$ in Eq.~\ref{eq:select dimension}; $\alpha =0.1$ in Eq.~\ref{eq_rectify propability}; and the iteration number $T$ in Algorithm~\ref{alg:scaled SBM} is empirically set to be 5.  The influence of some parameters will be investigated in details below, but the exact setting of these parameters is not critical: the method gives similar results over a broad range of settings.

\begin{figure} [!t]
\centering
\includegraphics[width=9cm]{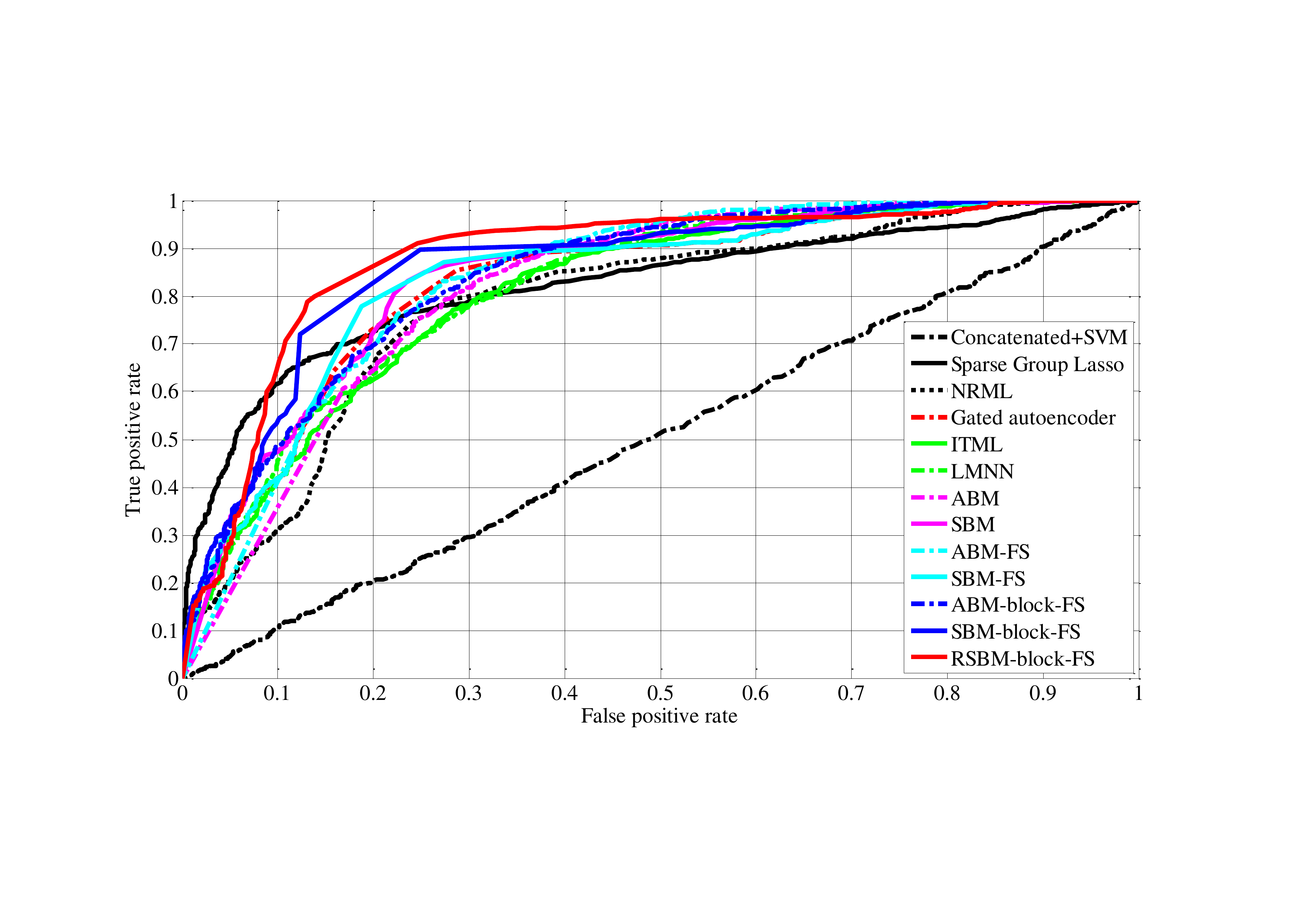}
\caption{The ROC curves of different methods obtained on the TSKinFace dataset.}
\label{fig:show ROC}
\end{figure}

\para{Comparison with the state of the art methods} Fig.~\ref{fig:show ROC} gives the ROC (Receiver Operating Characteristic) curves of different methods and Table~\ref{tab:compare results on LFKW for tri-objects} summarizes the results. One can see from the table that the performance of the baseline SVM algorithm gives an average accuracy of 53.4\%, indicating that the one-vs-two type tri-subject kinship verification is a very challenging problem. However, our proposed RSBM model working at the patch level improves this by over 30\%, being the best performer among all the compared methods. The closest competitor of our method is the DDML \cite{hu2014discriminative}, which gives an average accuracy of 81.0\% - similar to our method of `SBM-FS', but with the prior information exploited, our `RSBM-block-FS' performs better.

\begin{table} [!t]
\caption{Correct verification rates(\%) for different methods on the TSKinFace database (where ``FM-S",`FM-D" denote ``Father-Mother and Son" and ``Father-Mother and Daughter", respectively.)}
\begin{center}
\begin{tabular}{lccc}
\hline
Method & FM-S & FM-D & avg. \\
\hline
\hline
Concatenated+SVM   &53.5$\pm$0.2381   &53.2$\pm$0.2037    &53.4    \\
Sparse Group Lasso\cite{fangkinship} & 71.6$\pm$0.9644  & 69.8$\pm$0.3485  & 70.7 \\
NRML \cite{LuPAMI14}  &77.0$\pm$0.5831 & 71.4$\pm$0.5933 & 74.2 \\
Gated autoencoder \cite{dehghanlook}  &81.9 $\pm$0.4433 &79.6$\pm$0.3685 &80.8\\
DDML \cite{hu2014discriminative}  & 82.1 $\pm$1.0357  &79.8$\pm$0.5879 &81.0\\
\hline
ITML \cite{davis2007information}   &76.6$\pm$0.3753 & 71.4$\pm$0.4087 &74.0\\
LMNN \cite{weinberger2006distance} &75.4$\pm$0.7293  & 70.3$\pm$0.7372 &72.9\\
\hline
\hline
ABM (\emph{proposed}) & 78.5$\pm$0.3411 & 73.2$\pm$0.3888 & 75.9 \\
ABM-FS (\emph{proposed}) & 78.6$\pm$0.3114 & 76.9$\pm$0.2927 & 77.8 \\
ABM-block-FS (\emph{proposed}) & 83.4$\pm$0.2508 & 81.9$\pm$0.3025   &82.7    \\
\hline
SBM (\emph{proposed}) & 82.4$\pm$0.3568 & 78.2$\pm$0.4105 & 80.3 \\
SBM-FS (\emph{proposed}) & 82.8$\pm$0.2608 & 79.5$\pm$0.2550 & 81.2\\
SBM-block-FS (\emph{proposed}) & 85.2$\pm$0.3031 & 83.5$\pm$0.2985 &84.4 \\
\hline
RSBM-block-FS(\emph{proposed}) & \textbf{86.4$\pm$0.4105} & \textbf{84.4$\pm$0.3601} & \textbf{85.4}\\
\hline
\end{tabular}
\end{center}
\label{tab:compare results on LFKW for tri-objects}
\end{table}

Our method also significantly works better than the sparse group lasso based method proposed in Fang et al. \cite{fangkinship} - one possible explanation is that for a core family group involved only three subjects, the assumption made in \cite{fangkinship} that an image of a child should be best reconstructed by face images in his/her own family is too strong, although it is reasonable under their situation where dozens of face images per family are available.

Thirdly, we see that simply adopting state of the art metric learning methods for tri-subject kinship verification is not a good choice. This is partly due to the fact that these methods fail to model the dependence structure among the three visual entities. By contrast, the proposed RSBM model effectively exploits such priors during several stages of the verification pipeline (e.g.,similarity modeling, feature selection) and achieves the best verification performance.

Last but not least, it is interesting to note that the gender plays a significant role in kinship verification - in all the cases tested, the verification rates on ``FM-S" are consistently higher than those on ``FM-D". One possible reason is that the appearance variations appeared in a female subject (daughter) are more complex than those in a male subject (son). This seems to be in accordance with earlier psychological research results \cite{platek2004reactions} that the kin recognition signal is less evident from daughters than from sons.

%To further visualize the difference between our proposed methods and the other compared methods, the receiver operating characteristic curves of different methods are plotted in  Fig.~\ref{fig:show ROC}. We can see from this figure that the ROC curves of our prpoposed ``SBM-block w FS'' and ``RSBM-block w FS'' are higher than the other compared methods.

\begin{figure} [!t]
\centering
\includegraphics[width=9cm]{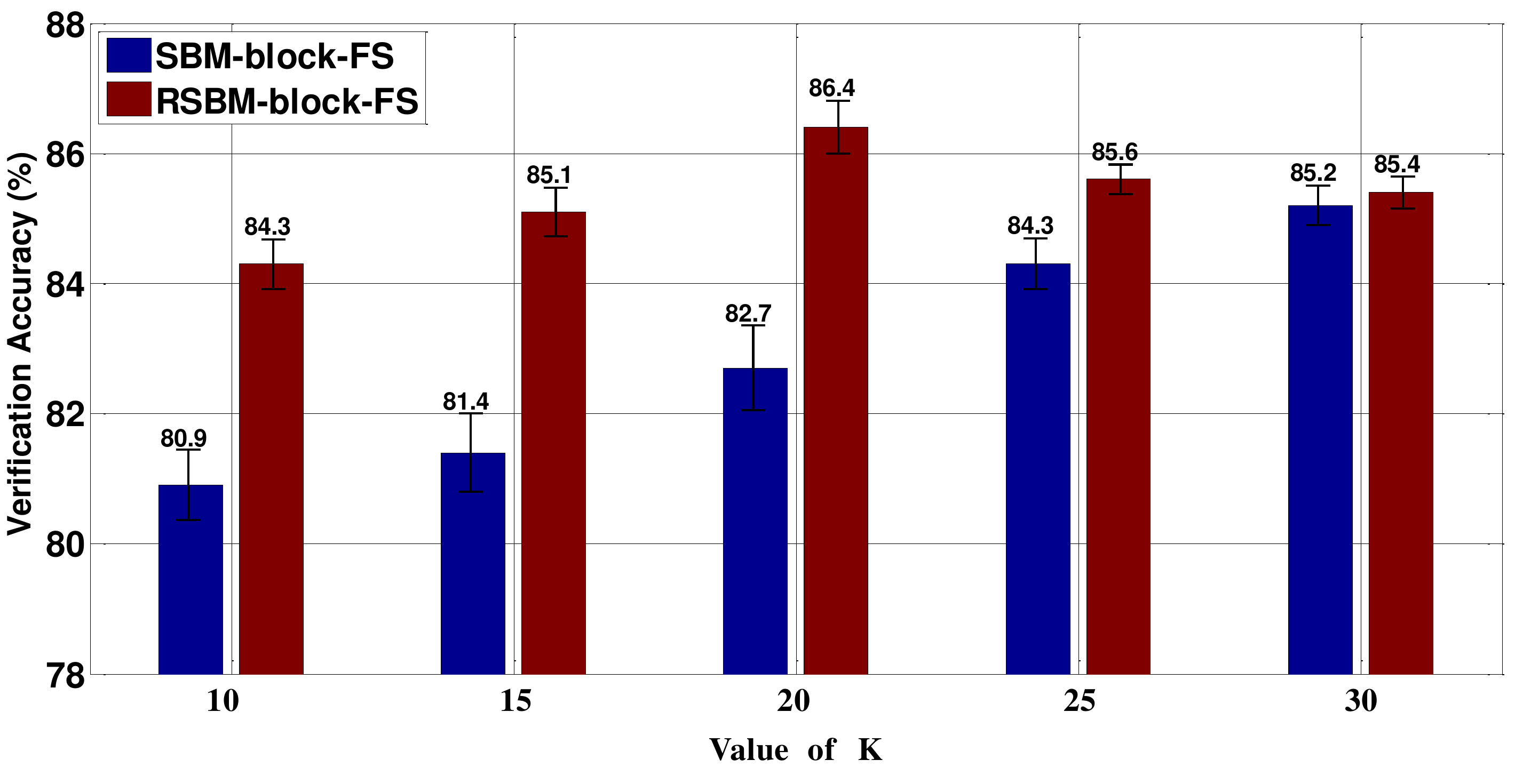}
\caption{Comparison of FM-S performance of SBM and RSBM, under different number $K$ of patches.}
\label{fig:show DSBM effect}\vspace{-2mm}
\end{figure}

\begin{figure} [!t]
\centering
\includegraphics[width=8cm]{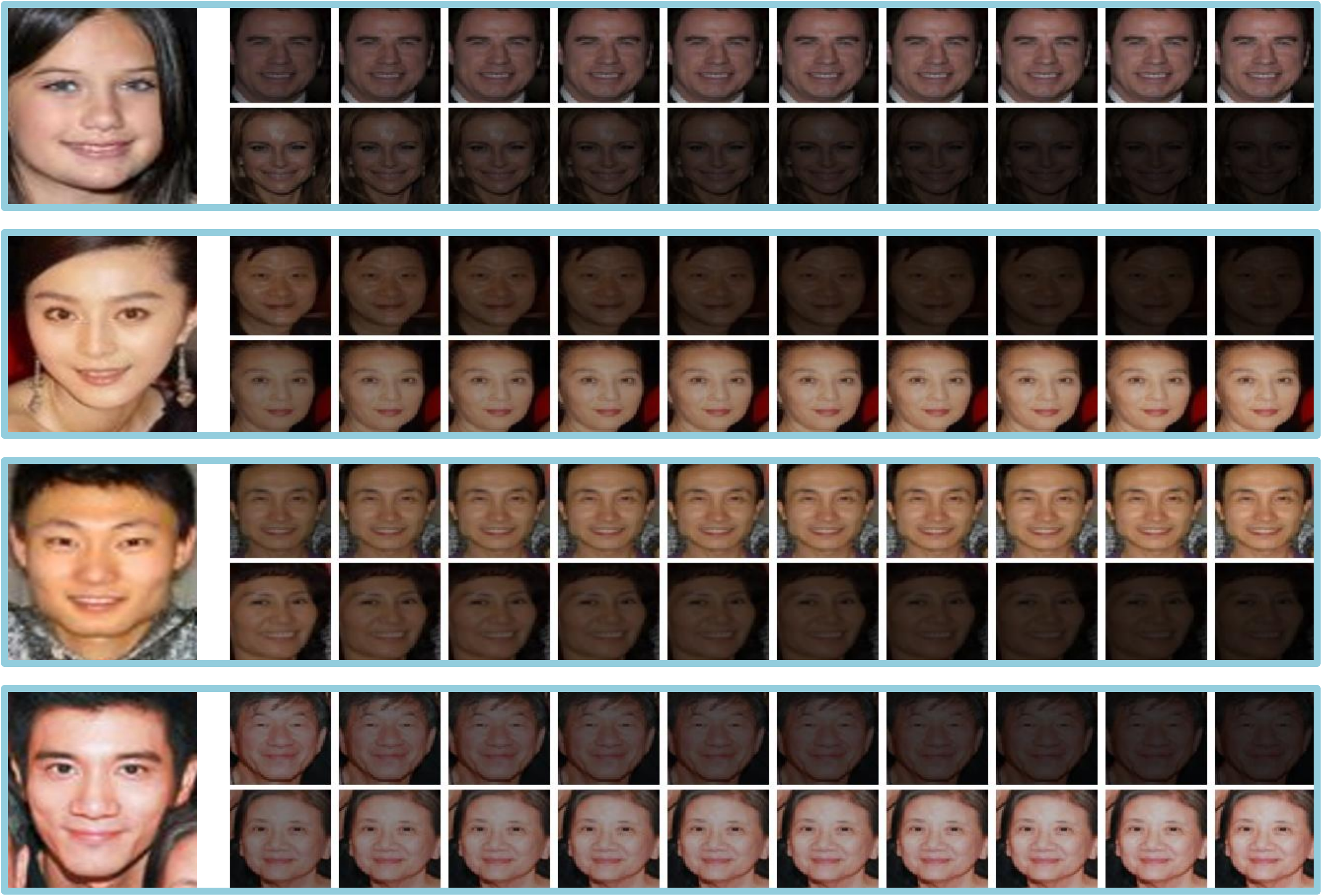}
\caption{Illustration of the learned prior knowledge for four families. In each family, the image on the left is the input child image, and the two rows of images on the right are the parents images multiplied by the respective learnt prior density in 10 iterations (progressively from left to right) - the higher the probability the lighter the pixel value. }
\label{fig:show propability}\vspace{-2mm}
\end{figure}

\para{The importance of prior knowledge} Fig.~\ref{fig:show DSBM effect} compares in detail the FM-S performance of the SBM model with/without exploiting the prior knowledge about the relative difference of a child to his/her parents, as a function of the number of patches selected for each face. One can see that when the number of patches selected for verification is relatively small, the RSBM method significantly performs better than the SBM method. For example, using only 20 patches, the accuracy of RSBM reaches an verification accuracy of 86.4\%, 3.7\% higher than that of the SBM model. This highlights the benefits of exploiting prior knowledge for complex kinship verification. To further illustrate this, we visualize the prior knowledge learned by multiplying it elementwise by the image: see Fig.~\ref{fig:show propability} for an example. We can observe that some children do look more like his/her father than mother,  or vice versa, and such information is effectively captured and utilized by our model.

\begin{figure}
\centering
\includegraphics[width=9cm]{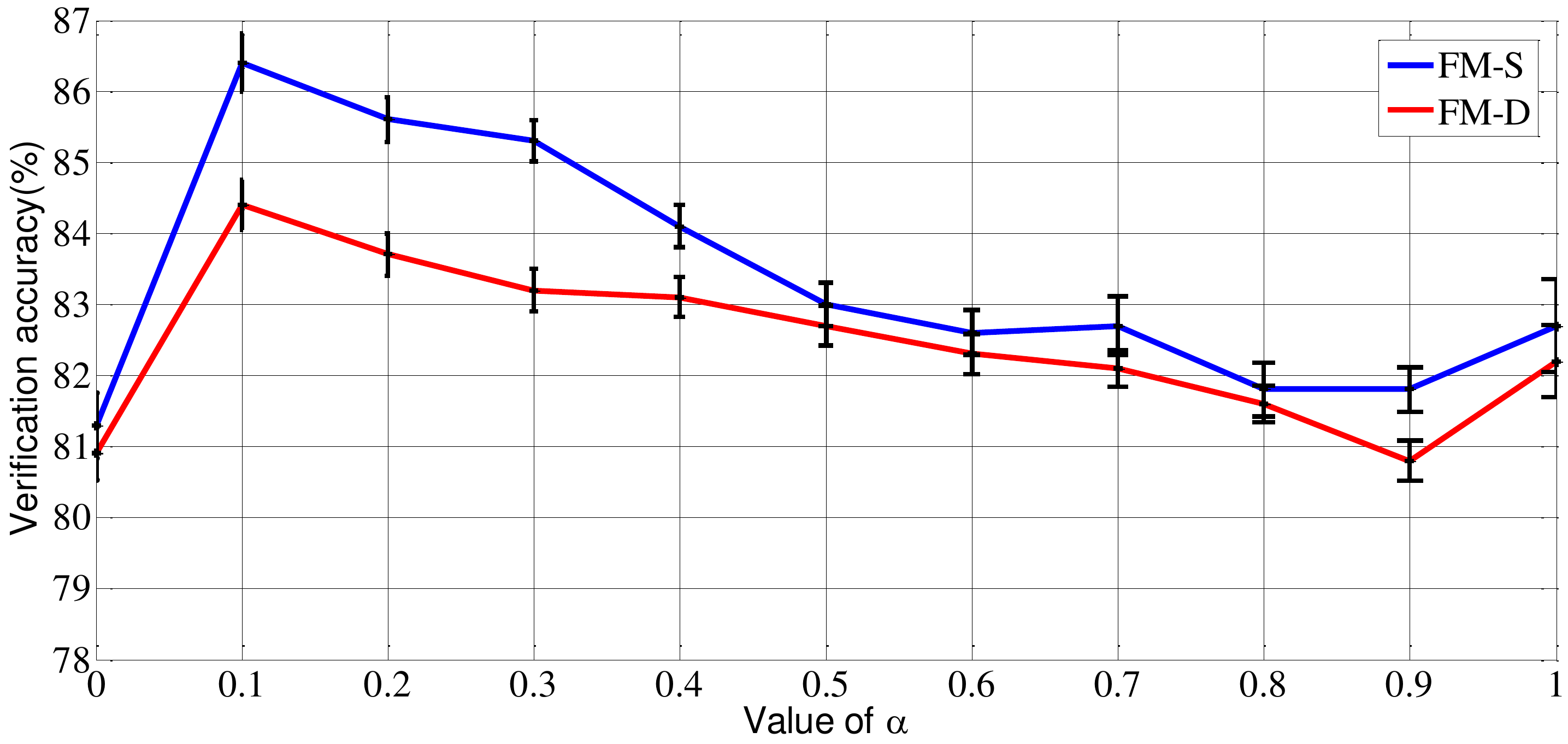}
\caption{The average performance of ``RSBM-block-FS'' as a function of the amount of stabilization $\alpha$.}
\label{fig:show alpha effect}\vspace{-2mm}
\end{figure}

Fig.~\ref{fig:show alpha effect}  shows the average verification accuracy of our RSBM model as a function of the stabilizing term $\alpha$ (c.f., Eq.~\ref{eq_rectify propability}). We can see that the RSBM model obtains the best performance when $\alpha = 0.1$ for both FM-S and FM-D. In general, good performance could be obtained by setting the value of $\alpha$ between $0.05$ and $0.3$.

\begin{figure}
\centering
\includegraphics[width=8cm]{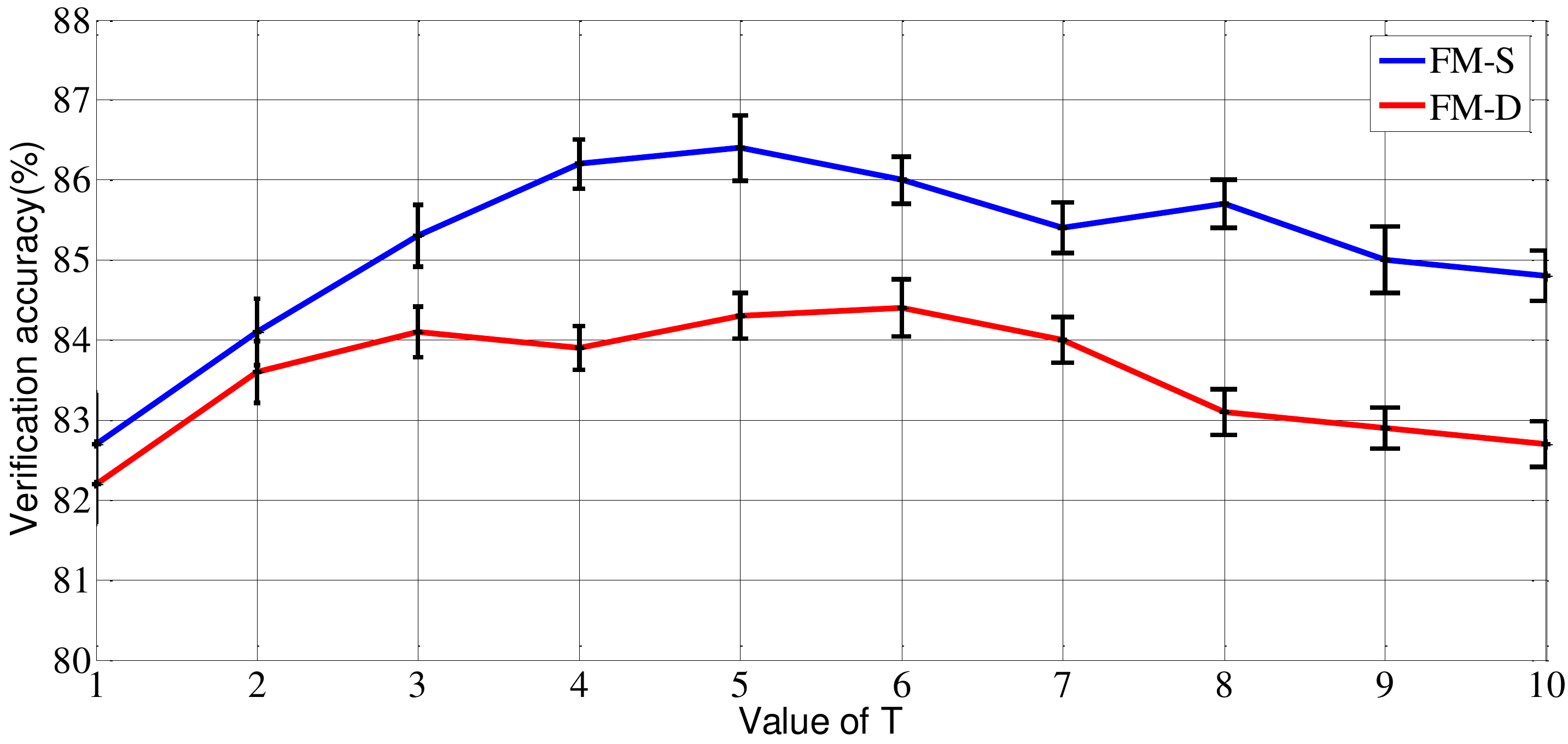}
\caption{The average verification accuracy of ``RSBM-block-FS'' as a function of the number of iteration T.}
\label{fig:show each iteration}\vspace{-2mm}
\end{figure}

Fig.~\ref{fig:show each iteration} shows the performance curve of the RSBM model as a function of the number of iterations. We can see that the performance of the RSBM model boosts to its highest value only after a few iterations. In practice we would recommend to set $T=5$ to avoid overfitting.

\para{Effectiveness of spatially voting for feature selection} To verify this, we compare our feature selection scheme with group lasso (GL) - both aim to automatically figure out a set of patches from face images for kinship verification. Particularly, we formulate the problem as the sparse group lasso penalized logistic regression in which the groups are defined as the patches. For a fair comparison, we set the parameter that controls the group weights as 0.88 for FM-D and 0.85 for FM-S, obtaining the same number of selected patches as that in our vote-based feature selection method. As the baseline we select the $l_1$ norm lasso algorithm, which performs the feature selection without using any spatial information.

%
%\begin{table}
%\caption{Correct verification rates(\%) for different feature selection methods on the TSKinFace database (where ``FS" denotes our vote based feature selection method while ``L1" denotes lasso and ``GL" denotes group lasso)}
%\begin{center}
%\begin{tabular}{llll}
%\hline
%Method & ~~~~FM-S &~~~~FM-D & avg. \\
%\hline
%\hline
%ABM-L1     &75.9$\pm$0.5100  &74.8$\pm$0.5716  &75.4    \\
%ABM-GL     & 76.0$\pm$0.4300   & 75.0$\pm$0.5523 & 75.5 \\
%ABM-FS (\emph{proposed}) & 78.6$\pm$0.3114 & 76.9$\pm$0.2927 & 77.8 \\
%\hline
%SBM-L1 & 78.3$\pm$0.4980 &79.2  $\pm$ 0.5629  &78.8      \\
%SBM-GL & 80.6$\pm$0.4972 & 80.9$\pm$0.5356 & 80.8 \\
%SBM-FS (\emph{proposed}) & 82.8$\pm$0.2608 & 79.5$\pm$0.2550 & 81.2\\
%\hline
%\end{tabular}
%\end{center}
%\label{tab:compare with group lasso}
%\end{table}

\begin{figure}
\centering
\includegraphics[width=9cm]{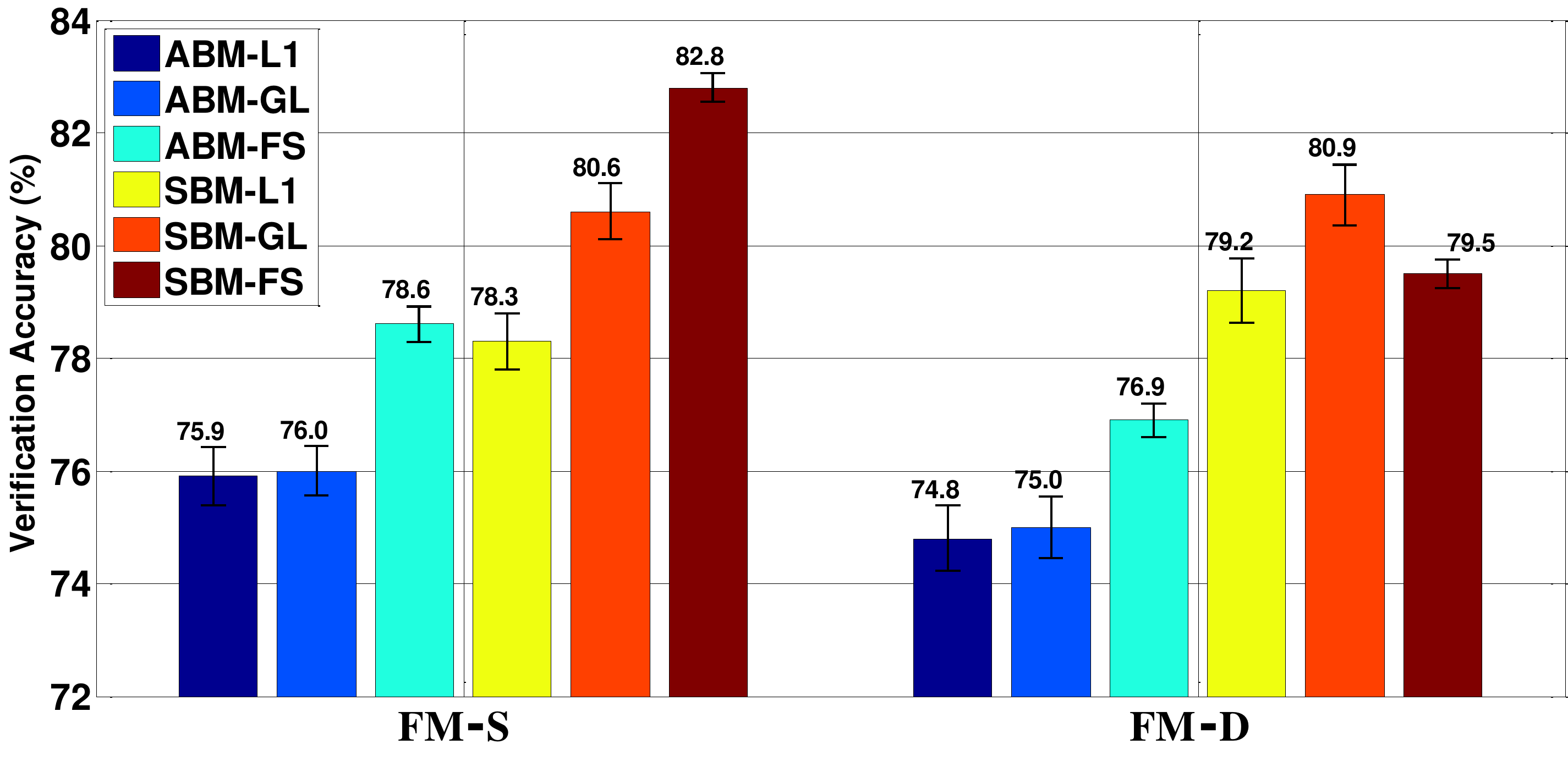}
\caption{Correct verification rates(\%) for different feature selection methods on the TSKinFace database (where ``FS" denotes our vote based feature selection method while ``L1" denotes lasso and ``GL" denotes group lasso)}
\label{fig:compare with group lasso}\vspace{-2mm}
\end{figure}

Figure~\ref{fig:compare with group lasso} gives the results. One can see that the proposed spatially voted feature selection scheme (``FS") performs better than the group lasso ``GL", on average improving the performance by about 2.3\% and 0.4\%, respectively on both tasks, while the simple lasso method performs the worst.

%We are also interested in the relative importance of patches in a face image evaluated by our feature selection method. For this, we use the votes received by each patch as its weight, according to which the patch is plotted with different degree of lightness. Fig.~\ref{fig:show selected patch} shows the images from four families with weight information imposed. The figure reveals that the most important patches seem to be distributed around the center of his/her face and hair area and some other edge regions are not very important. Actually, those important pathes contain facial organs about one's face. And we can observe that the nose and mouth are more discriminative than eye and eyebrows regions. Hence they are useful to be utilized to enhance the generalization ability of our verification model. Noting that these regions are mined from only a few hundreds of face groups in our database, it is still desirable to collect even larger family-based kinship database to further exploit such discriminative information from facial patches.
%
%\begin{figure}[!htb]
%\centering
%\includegraphics[width=9cm]{p8}
%\caption{Illustration of the effects of feature selection. Four groups of the images of a father, a mother and their child are shown, with the lightness of patches on each face positively correlated to the number of votes received by it. }
%\label{fig:show selected patch}
%\end{figure}
%

\begin{figure}[!t]
\centering
\includegraphics[width=8cm]{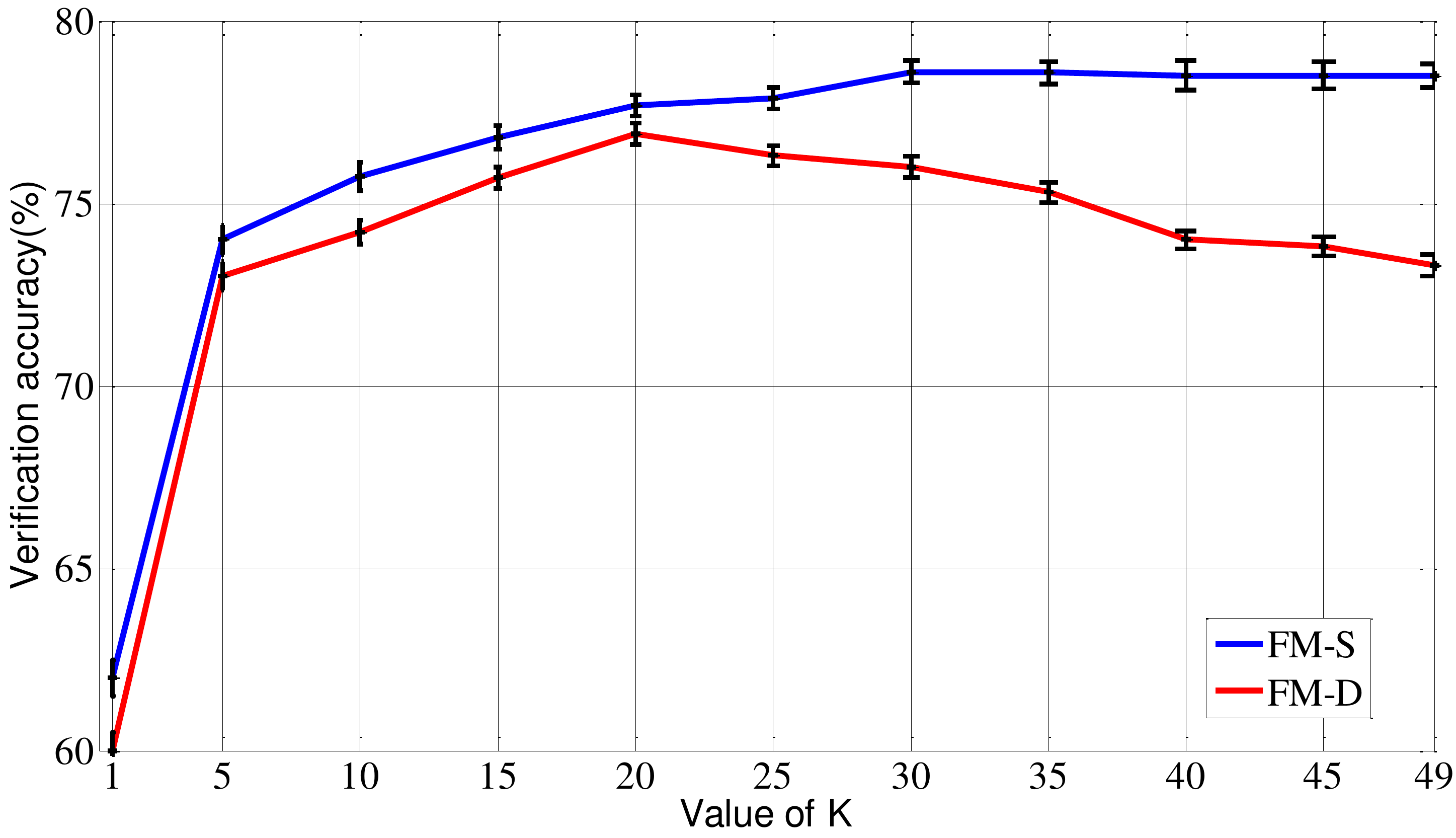}
\caption{Influence of the number of patches K selected on the verification rates.}
\label{fig:accuracy of different B}\vspace{-4mm}
\end{figure}

To answer the question of how many patches it needs to be selected, we investigate the effect of the parameter $K$ (the number of patches selected) on the performance. Fig.~\ref{fig:accuracy of different B} shows verification rate as a function of the number of patches selected for each face, with the ABM model as the verifier. One can see that the performance boots from about 60.0\% to over 73.0\% with only 5 patches. The performance increases with more patches added until 20 patches are selected, and the improvement is not evident after that for the FM-S verification. While for the FM-D verification, the number of selected patches is better to be kept less than 20 so as to reduce the possible influence of noise.

\begin{table*} [!t]
\caption{Correct rates (\%) of different methods for bi-subject kinship verification with triple inputs(column 2 and 5) and pair inputs (column 3,4,6 and 7, ``*'' denotes that the result (P-values) of $t-$test for the performance comparison between pair inputs and triple inputs verification is less than 0.05).}
\begin{center}
\begin{tabular}{l|lll|lll}
\hline
Method & ~~~~FM-S & ~~~~FS & ~~~~MS & ~~~~FM-D & ~~~~FD & ~~~~MD \\
\hline
\hline
Sparse Group Lasso\cite{fangkinship} &71.6$\pm$0.9644 &69.1$\pm$0.6093   &68.7$\pm$1.2204  &69.8$\pm$0.3485 &66.8$\pm$0.4627(*)  &67.9$\pm$0.5977  \\
NRML \cite{LuPAMI14} & 77.0$\pm$0.5831 & 74.8$\pm$0.7279(*) & 72.2$\pm$0.3360(*) &71.4$\pm$0.5933 &70.0$\pm$0.6716(*) & 71.3$\pm$0.5853 \\
Gated autoencoder \cite{dehghanlook}  &81.9$\pm$0.4433 &79.9$\pm$0.6790(*)  &78.5$\pm$0.5963(*) &79.6$\pm$0.3686 &74.2$\pm$0.3170(*)  &76.3$\pm$0.2296(*)  \\
\hline
ITML \cite{davis2007information}  &76.6$\pm$0.3753  &75.6$\pm$0.3866(*)  &72.1$\pm$0.3330(*)  &71.4$\pm$0.4087  &70.5$\pm$0.4000(*)  &70.7$\pm$0.4435(*)  \\
LMNN \cite{weinberger2006distance} &75.4$\pm$0.7293  &72.7$\pm$0.7305  &71.5$\pm$0.7455(*) &70.3$\pm$0.7372 &69.8$\pm$0.7243(*)  &70.1$\pm$0.3846  \\
\hline
\hline
ABM-block-FS (\emph{proposed})  & 83.4$\pm$0.2508   & 83.0$\pm$0.5558  & 82.8$\pm$0.5037  & 81.9$\pm$0.3025   & 80.5$\pm$0.4301  & 81.1$\pm$0.4003\\
SBM-block-FS (\emph{proposed}) & 85.2$\pm$0.3031 & 83.0$\pm$0.5558(*)   & 82.8$\pm$0.5037(*)   & 83.5$\pm$0.2985 & 80.5$\pm$0.4301(*)  & 81.1$\pm$0.4003(*)\\
RSBM-block-FS(\emph{proposed})&\textbf{86.4$\pm$0.4105}  &83.0$\pm$0.5558(*)  &82.8$\pm$0.5037(*) &\textbf{84.4$\pm$0.3601} &80.5$\pm$0.4301(*) &81.1$\pm$0.4003(*) \\
\hline
\end{tabular}
\end{center}
\label{tab:compare results on LFKW}
\end{table*}

\begin{table*} [!t]
\caption{Correct verification rates(\%) for different methods on the TSKinFace database(where "FS-M", "MS-F", "FD-M" and "MD-F" denote "father-son and mother", "mother-son and father", "father-daughter and mother" and "mother-daughter and father", respectively.))}
\begin{center}
\begin{tabular}{lccccc}
\hline
Method & FS-M  &  MS-F   & FD-M    &  MD-F   &avg.\\
\hline
\hline
Concatenated+SVM      &53.5$\pm$0.2066   &53.8$\pm$0.1337  &53.0$\pm$0.1732   &53.3$\pm$0.2523   &53.4   \\
Sparse Group Lasso\cite{fangkinship}     &69.9$\pm$0.4927   & 70.3$\pm$0.7489   &68.6$\pm$0.5350   &68.9$\pm$0.5070    &69.4      \\
NRML \cite{LuPAMI14}     &75.6$\pm$0.5931   &75.8$\pm$0.3788   &70.8$\pm$0.9266   &70.2$\pm$0.3189   &73.1      \\
Gated autoencoder \cite{dehghanlook}     &79.7$\pm$0.4077   &80.9$\pm$0.4023     &78.4$\pm$0.4022   &79.2$\pm$0.4133  &79.6   \\
\hline
ITML \cite{davis2007information}   &73.8$\pm$0.4921   &74.6$\pm$0.2564   &70.1$\pm$0.3727   &70.0$\pm$0.4950  &72.1      \\
LMNN \cite{weinberger2006distance} &71.9$\pm$0.2372   &72.6$\pm$0.7213   &70.5$\pm$0.5129   &69.5$\pm$0.4232  &71.1            \\
\hline
\hline
ABM  (\emph{proposed})    &78.0$\pm$0.4379   &78.7$\pm$0.3117   &73.1$\pm$0.4190   &73.5$\pm$0.5103 &75.8     \\
ABM-FS (\emph{proposed})     &77.9$\pm$0.3104    &78.0$\pm$0.4235   &75.1$\pm$0.3008   &76.5$\pm$0.4730  &76.9     \\
ABM-block-FS (\emph{proposed})    &81.3$\pm$0.2791  &81.8$\pm$0.3984   &80.4$\pm$0.3604   &80.7$\pm$0.3309  &81.1        \\
\hline
SBM (\emph{proposed})    &79.2$\pm$0.4113    &80.4$\pm$0.4103   &76.8$\pm$0.4216   &77.1$\pm$0.4010  &78.4      \\
SBM-FS (\emph{proposed})    &81.0$\pm$0.3716    &81.7$\pm$0.3002   &78.7$\pm$0.3919   &79.0$\pm$0.2637  &80.1       \\
SBM-block-FS (\emph{proposed})    &\textbf{82.9$\pm$0.1841}    &\textbf{83.9$\pm$0.4197}  &\textbf{81.9$\pm$0.1071}   &\textbf{81.8$\pm$0.2157}  &82.6      \\
\hline
\end{tabular}
\end{center}
\label{tab:compare results on Different}
\end{table*}

\begin{figure}[!t]
\centering
\includegraphics[width=7cm]{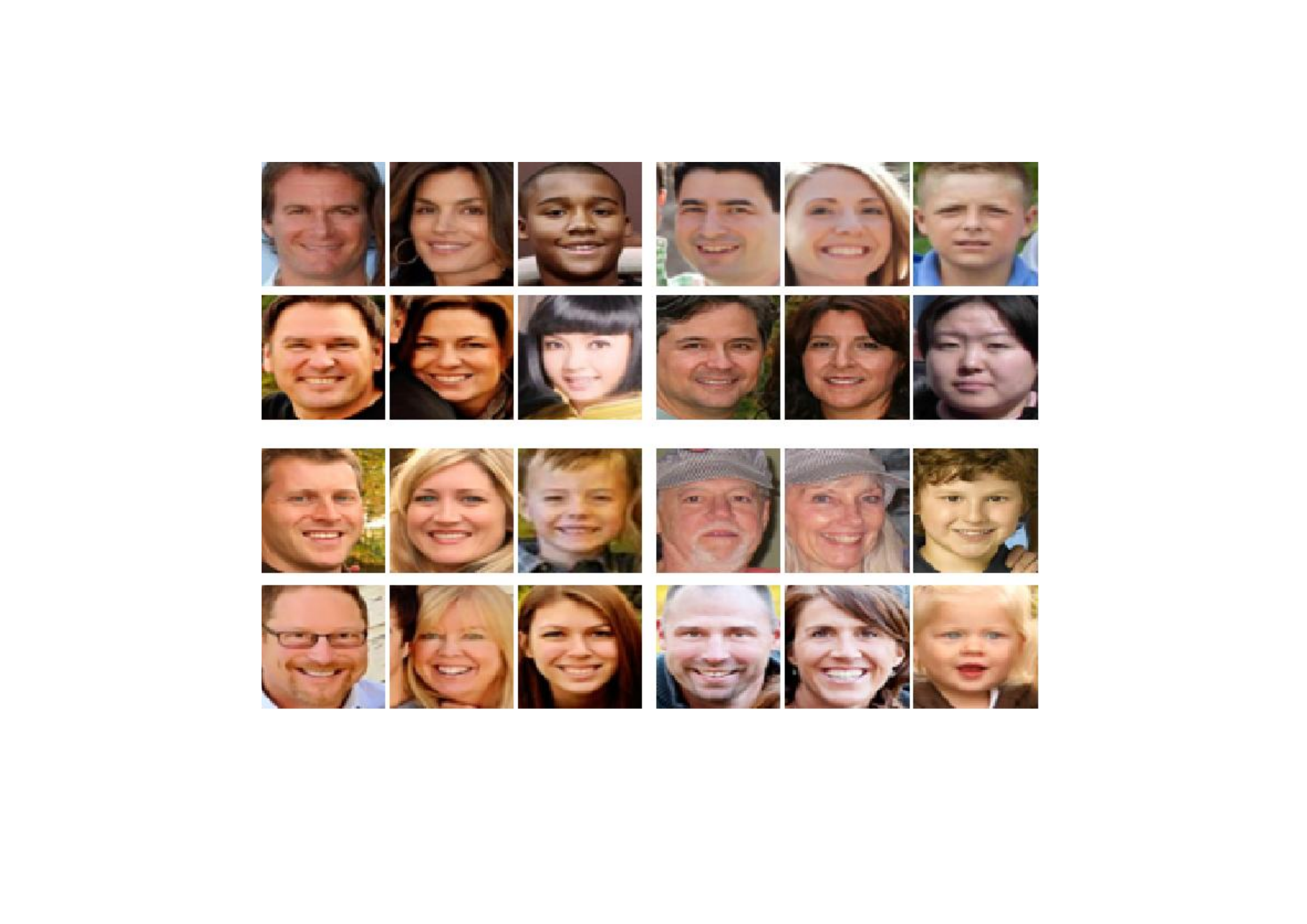}
\caption{Illustration of some samples in the easy negative set (the top two rows) and the hard negative set (the bottom two rows). }
\label{fig:kmeans on easy and hard}\vspace{-4mm}
\end{figure}

\para{Influence of randomness in negative sample generations} In the previous experiments the negative samples are randomly generated by combining child and parents from different families but there is only one for each. We now investigate the impact of such randomness. Particularly, we first randomly generate a large negative samples pool containing both the FM-D and the FM-S negative relationship. Although there are many ways to distinguish less distinct negative samples (i.e., hard samples with features of child and parents not very different) from those very distinct negative samples (i.e, easy ones with features of child and parents quite different), for example, by simply measuring the similarity between the child and parents, we choose the ``LMNN'' method \cite{weinberger2006distance} here. That is, those samples correctly classified by the ``LMNN'' method as negative are categorized as ``easy'' ones, otherwise as ``hard''. We use this criterion to randomly select 2,000 samples from the negative pool, with 1,000 each for the ``easy'' and the ``hard'' category respectively. Some of these samples are illustrated in Fig.~\ref{fig:kmeans on easy and hard}. For experiment we equally split those samples into 10 sets, and run the model trained on the fifth fold of our TSKinFace database over them.

Table ~\ref{tab:accuracy of nega test} gives the results. One can see that all the methods investigated here work consistently and significantly better on the easy set than on the hard set, indicating that it makes sense to distinguish the two sets based on the outputs of the ``LMNN'' method, while our ``RSBM-block-FS'' method demonstrates the highest robustness against this random confusion. The table also reveals that although some of the methods work well on the easy negative samples, the performance on the hard ones are not satisfactory in general (with accuracy less than 70.0\%), showing that further research is needed on this topic.

\begin{table} [!t]
\scriptsize
\caption{Correct verification rates(\%) for different methods on the 5-th fold of TSKinFace database with 10 different negative samples sets, where 'Easy' represents those samples which are classified by ``LMNN'' correctly and 'Hard' denotes those samples which are classified by ``LMNN'' incorrectly}
\begin{center}
\begin{tabular}{lccc}
\hline
Method  & Easy  & Hard  &avg.\\
\hline
\hline
LMNN \cite{weinberger2006distance}   &100.0      &0.0   &50     \\
Concatenated+SVM  &46.3$\pm$0.0224       &38.1$\pm$0.0173   &42.2$\pm$0.0176       \\
NRML \cite{LuPAMI14} &78.4$\pm$0.0046           &51.4$\pm$0.0256   &64.9$\pm$0.0053        \\
DDML \cite{hu2014discriminative} &77.6$\pm$0.0083           &55.1$\pm$0.0163   & 66.4$\pm$0.0086   \\
ABM-block-FS (\emph{proposed})   &87.3$\pm$0.0041          &54.0$\pm$0.0146   &70.7$\pm$0.0030  \\
SBM-block-FS (\emph{proposed})  &92.5$\pm$0.0037           &57.3$\pm$0.0143   &74.9$\pm$0.0036    \\
RSBM-block-FS (\emph{proposed}) &\textbf{94.7$\pm$0.0040}          &\textbf{63.1$\pm$0.0142}   &\textbf{78.9$\pm$0.0032}   \\
\hline
\end{tabular}
\end{center}
\label{tab:accuracy of nega test}
\end{table}

\para{Computational Complexity} We now briefly analyze the computational complexity of the RSBM method, which involves T iterations. In each iteration we solve a regularized logistic regression problem and make the estimation of the weights of $p^{fc}$ and $p^{mc}$. To solve the regularized logistic regression problem, we use a fast implementation \cite{ji2009accelerated} with its computational complexity O(${d^3N/g^2}$), where $g$ is the iteration counter, while the computational complexity of the estimating part is $O(N)$. Hence the total computational complexity of our proposed RSBM is O(${d^3N/g^2}T$)+O(NT).

\subsection{Enhancing Bi-Subject Kinship Verification}\label{sec_enhance bi-subject}
Intuitively, having more information about one's parents is potentially useful to improve the performance of bi-subject kinship verification. In order to verify this hypothesis, another series of experiments are conducted. This is similar to the traditional bi-subject verification in that four types of kinship relations will be evaluated, i.e., Father-Son(FS), Father-Daughter (FD), Mother-Son (MS), and Mother-Daughter (MD). However, the key difference lies in that we are now  given a triple including two parents and a child as a test sample. In other words, we are interested in, for example, whether the information about one's father is useful to verify the Mother-Daughter (MD) relation.

\begin{figure} [!t]
\centering
\includegraphics[width=6cm]{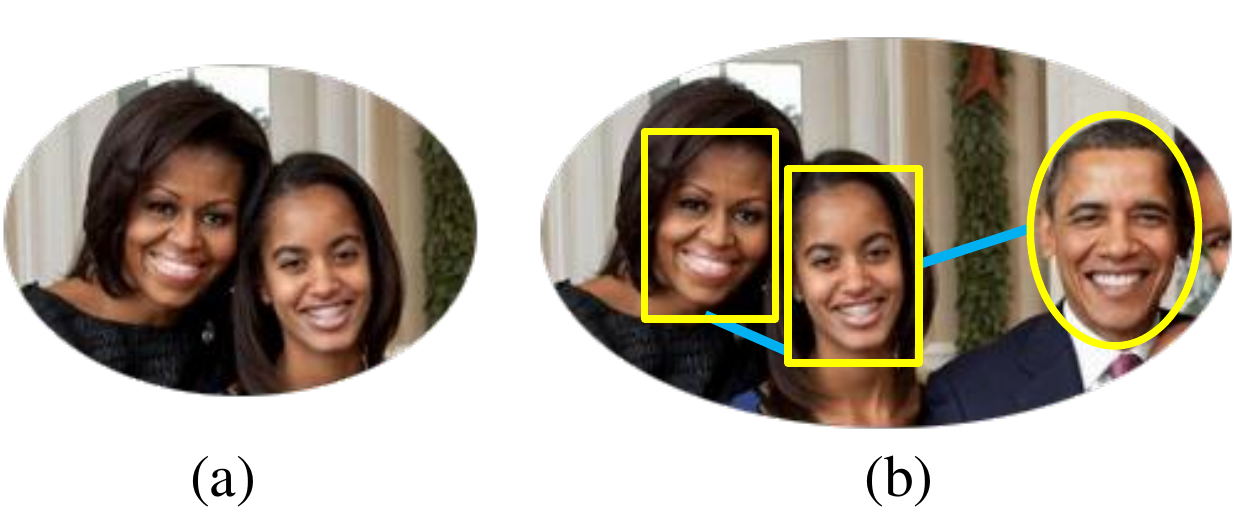}
\caption{When the images of a second parent is available, the traditional one-vs-one type bi-subject kin verification problem can be reformulated as a one-vs-two one. In this example, once the Father/Mother-Daughter (FM-D) relationship is established for the three subjects shown on the right, one can safely infer that the Mother-Daughter (MD) kinship is validated for the subjects shown on the left.
}
\label{fig:show tri effect}
\end{figure}

One simple way for this is to reformulate the bi-subject kinship verification problem as a tri-subject problem, since once a FM-D relationship is established, a FD and a MD relationship \emph{must} be established as well, see Fig.~\ref{fig:show tri effect} for an example. For a bi-subject verification problem shown in Fig.~\ref{fig:show tri effect}(a), one can treat it as a tri-subject problem shown in Fig.~\ref{fig:show tri effect}(b) when the father's information is available, and use that result to answer the question of one-vs-one kinship verification.

Table~\ref{tab:compare results on LFKW} compares the results of these two approaches for bi-subject kinship verification. One can see that exploiting more information about one's parents is indeed beneficial. Particularly, the performance of the mother-son (MS) verification is improved significantly from 72.2\% to 77.0\% using the SVM-based NRML baseline method, while that of the father-daughter (FD) verification is improved from 80.5\% to 84.4\% using our  RSBM model, and $t$-test analysis shows that this improvement is statistically significant. Particularly, the stars in Table~\ref{tab:compare results on LFKW} indicate whether the improvement of performance for triple inputs is significant compared to that when only two subjects are available. For example, the accuracy of 'SBM-block-FS' for father-son verification (FS) is $83.0\%$, but if the information about the mother is known, this improves to $85.2\%$ (as shown in the column of 'FM-S'), which is significantly better than that of FS according to the t-Test and hence a star is marked, otherwise there is no star.

It is well known that a problem like MS or FD verification is quite difficult due to the different genders of two subjects to be verified. Our method essentially provides a new solution to this, and we consider it as one of the major motivations to study the tri-subject kinship verification problem.

\subsection{Comparisons with Human Beings in Kinship Verification}

To investigate human beings' performance on the kinship verification problem, we randomly selected 100 groups of cropped grayscale face samples, including 50 positive groups and 50 negative ones. Then we presented these to 10 human observers with ages of 20 to 40 years old to ask their opinions about the kinship relation. These human observers did not receive any training on how to verify kinship from facial images before the experiment, and will completely rely on their own knowledge accumulated to answer the questions.

Particularly, we conduct two parts of tests on kinship verification. For the first part, 100 child-parent pairs (one-vs-one) are shown to human observers (``A'' ), and for the second part, 100 child-parents groups (one-vs-two) are presented to these observers (``B'' ). Obviously, these two types of testing are respectively corresponding to the problem of bi-subject and tri-subject kinship verification. We repeated this procedure two times, one for the FM-S subset and other for the FM-D subset, both from our TSKinFace database. We also run the same experiments using our SBM method for comparison.

Table ~\ref{tab:Human_test_FMS} and Table ~\ref{tab:Human_test_FMD} give the results. One can see that ``B'' can obtain better performance than ``A'' on the two subsets, which indicates that human beings are able to combine the information from both parents to make better kinship judgement. For example, the performance of the mother-son( MS ) verification is 74.2\%,  but if the face image of one's father is also available, the performance increases to 79.9\%. Moreover, it is worth mentioning that our proposed SBM methods achieve higher verification accuracies than ``B''.

\begin{table} [!b]
\scriptsize
\caption{Correct rates (\%) of human beings and our method on the FM-S subset of the TSKinFace database.}
\begin{center}
\begin{tabular}{l|ccc}
\hline
 Method &FM-S  &FS  &MS     \\
\hline
\hline
 A &N/A              &77.3$\pm$2.1927   &74.2$\pm$1.6592    \\
 B &79.9$\pm$1.6362   &N/A            & N/A                  \\
\hline
\hline
SBM(\emph{proposed})  &81.6$\pm$0.2875    &N/A    &N/A     \\
SBM-FS(\emph{proposed})  &81.9$\pm$0.3479               &N/A    &N/A     \\
SBM-block-FS(\emph{proposed})  &82.4$\pm$0.6419               &N/A    &N/A     \\
\hline
RSBM-block-FS(\emph{proposed}) &85.4$\pm$0.1789               &N/A    &N/A     \\
\hline
\end{tabular}
\end{center}
\label{tab:Human_test_FMS}
\end{table}

\begin{table} [!b]
\scriptsize
\caption{Correct rates (\%) of human beings and our method on the FM-D subset of the TSKinFace database.}
\begin{center}
\begin{tabular}{l|ccc}
\hline
 Method  &FM-D   &FD &MD    \\
\hline
\hline
 A     &N/A            &73.5$\pm$1.2042   &75.5$\pm$1.2942 \\
 B     &79.2$\pm$1.4415   &N/A           &N/A      \\
\hline
\hline
SBM(\emph{proposed})  &79.2$\pm$0.4131               &N/A    &N/A     \\
SBM-FS(\emph{proposed})  &80.0$\pm$0.6127               &N/A    &N/A     \\
SBM-block-FS(\emph{proposed})  &81.4$\pm$1.0354               &N/A    &N/A     \\
\hline
RSBM-block-FS(\emph{proposed}) &83.0$\pm$0.8000               &N/A    &N/A     \\
\hline
\end{tabular}
\end{center}
\label{tab:Human_test_FMD}
\end{table}

\subsection{Robustness under Different Lighting Conditions}
Since all the face images in a family in our database are extracted from the same photo, it could introduce unnecessary bias in learning. In order to investigate the behavior of our algorithm when encounters face images from completely different lighting conditions, we construct a new dataset based on the Family101 \cite{fangkinship}.

\begin{figure}[!t]
\centering
\includegraphics[width=6cm]{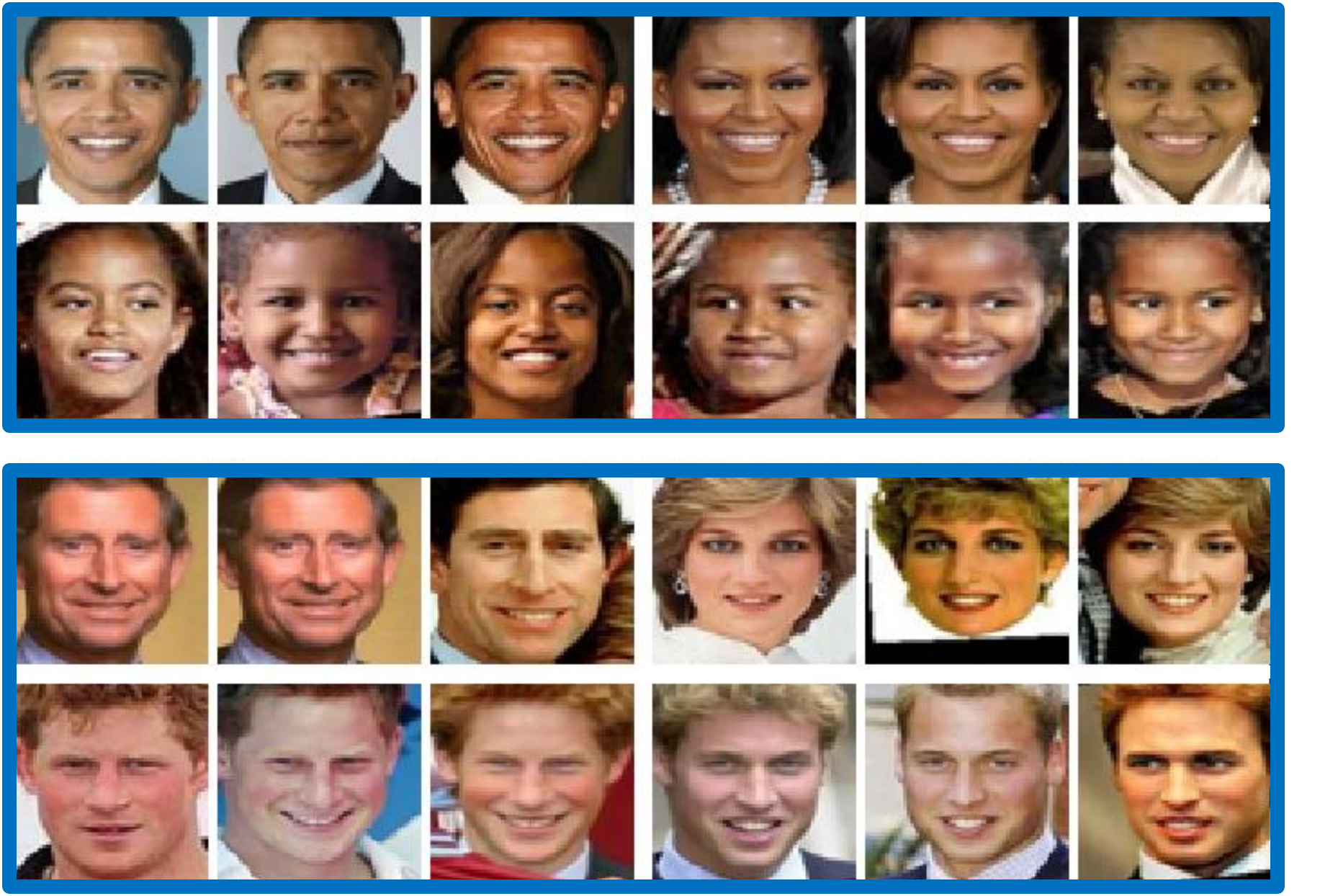}
\caption{Illustration of the construction of a new test dataset with different lighting conditions using the face images from the Family101 database. For each family, there are several face images per subject (upper row: parents, lower row: two children). We randomly select one image from two parents and one child to construct a triple-item group, such that all the images from the same group do not come from the same photo.}
\label{fig:samples from Family101}
\end{figure}

Particularly, we manually selected 48 families from 206 nuclear families of Family101, with the following conditions: 1) each family contains four members, i.e., father, mother and two children and 2) at least 3 face images exist for each family member. We then cropped these images to $64 \times 64$ pixels and converted them to 8 bit gray-scale, divided them into  $7 \times 7$ overlapping patches and extracted SIFT features. Fig.~\ref{fig:samples from Family101} gives some examples of the preprocessed images.

%{\color{blue}{For evaluation, we adopt a four-fold cross validation protocol similar to previous experimental settings: equally dividing the database into four folds such that each fold contains nearly the same number of face groups. For each fold, there is 12 families, with each family 3 father images and 3 mother images. First, we construct 9 father-mother pairs. Then each father-mother pair can be combined with 6 child images from this family for positive groups. Thus, we can construct 54 groups for each training family. Thus, $54\times 12\times3=1944$ positive groups can be obtained without repetition and no any three face images in each group appear in one photos before. This will suppress the bias for the positive samples to have similar lighting conditions as much as possible.}}

Then, to construct tri-subject groups for our tri-subject kinship verification, we do the following iterations for each of the selected family:
\begin{enumerate}
  \item select one image among 3 images from the father's.
  \item select one image among 3 images from the mother's.
  \item select one image among 6 images from the two children's.
\end{enumerate}

This will give us $3\times 3\times 6 = 54$ different groups per family. In the experiments,  we follows the four-fold cross validation protocol, which means that for each round of evaluation, 36 families will be used for training while the remaining 12 for testing. In other words, in training we will have 36 families $\times$ 54 groups/family = 1944 groups in total. And no any three face images in each group appear in one photos before. This will suppress the bias for the positive samples to have similar lighting conditions as much as possible.

\begin{figure}
\centering
\includegraphics[width=9cm]{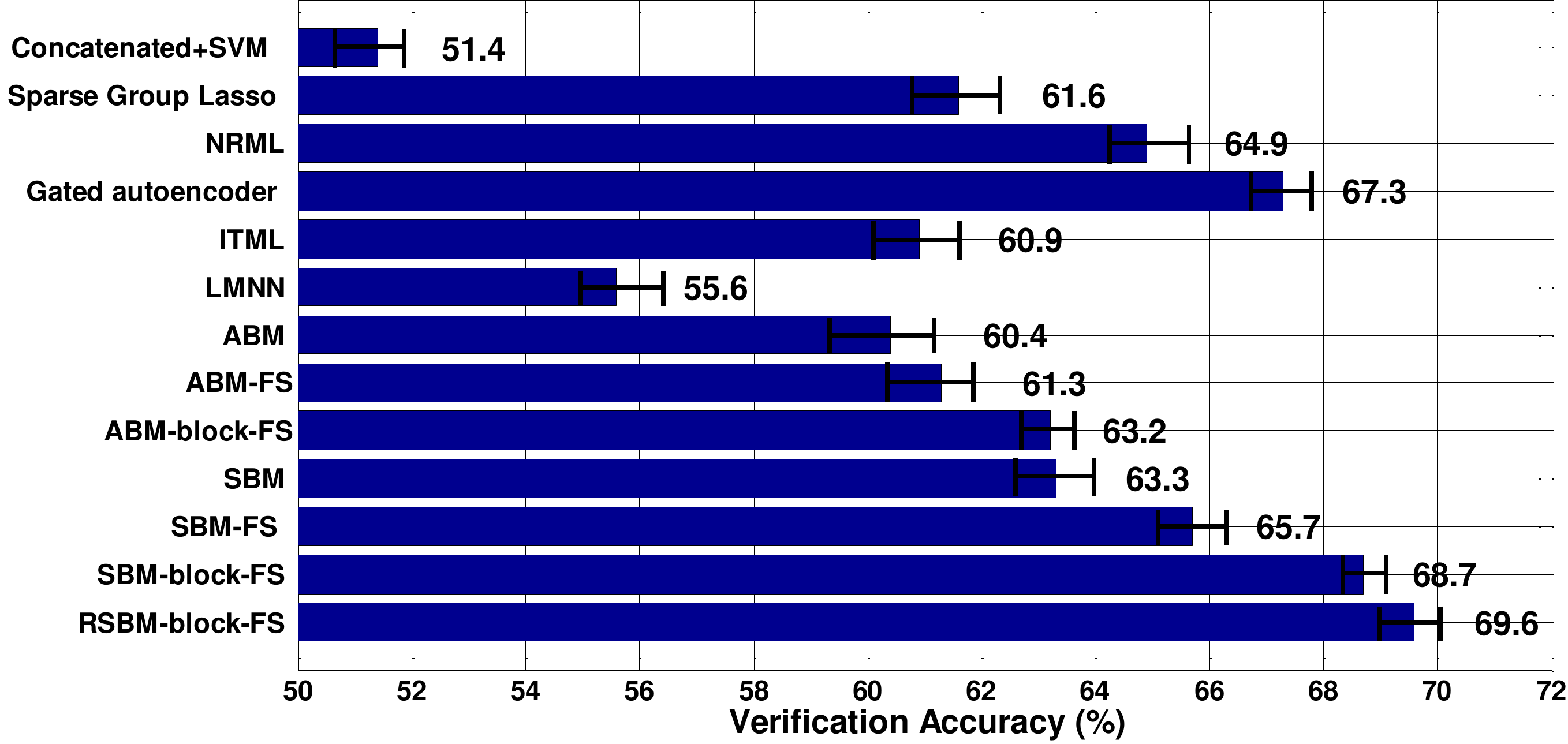}
\caption{Correct verification rates(\%) for different methods on the Family101 subset database}
\label{fig:compare results on Family101}\vspace{-2mm}
\end{figure}

Figure~\ref{fig:compare results on Family101} gives the results. One can see that all the methods are influenced by the illumination changes introduced in the dataset. However, the proposed `RSBM-block-FS' performs the best among the compared ones, about 18.2\% higher than the baseline algorithm in terms of accuracy. The table also reveals that by replacing the pairwise bilinear similarity with the proposed relative similarity measure, one can improve the performance from 68.7\% (`SBM-block-FS') to 69.6\% (`RSBM-block-FS').

%\begin{table} [!t]
%\caption{Correct verification rates(\%) for different methods on the Family101 subset database)}
%\begin{center}
%\begin{tabular}{lccc}
%\hline
%Method & FM-C   \\
%\hline
%\hline
%Concatenated+SVM      &51.4$\pm$0.5221       \\
%Sparse Group Lasso\cite{fangkinship}     &61.6$\pm$0.8733       \\
%NRML \cite{LuPAMI14}     &64.9$\pm$0.7619     \\
%Gated autoencoder \cite{dehghanlook}     &67.3$\pm$0.4423    \\
%\hline
%ITML \cite{davis2007information}   &60.9$\pm$0.7938      \\
%LMNN \cite{weinberger2006distance} &55.6$\pm$0.5795    \\
%\hline
%\hline
%ABM  (\emph{proposed})    &60.4$\pm$1.0432     \\
%ABM-FS (\emph{proposed})     &61.3$\pm$0.8325     \\
%ABM-block-FS (\emph{proposed})    &63.2$\pm$0.3310       \\
%\hline
%SBM (\emph{proposed})    &63.3$\pm$0.6381      \\
%SBM-FS (\emph{proposed})    &65.7$\pm$0.5319        \\
%SBM-block-FS (\emph{proposed})    &68.7$\pm$0.2777      \\
%\hline
%RSBM-block-FS(\emph{proposed})   & \textbf{79.6$\pm$1.0366}    \\
%\hline
%\end{tabular}
%\end{center}
%\label{tab:compare results on Family101}
%\end{table}

%It can be seen that the results obtained on Family101 subset are generally lower than those obtained on our database in Table~\ref{tab:compare results on LFKW for tri-objects}, which indicates that the kinship verification on Family101 is more difficult than on our database. The reason is that face images in our database are collected from the same photo and the kinship images have the same collection conditions, which could reduce some challenges caused by the illumination in the Family101 subset.
%
%

\subsection{Other Forms of Tri-Subject Kinship Verification}

In previous sections we focus on the child-parents type tri-subject kinship verificaiton, but the same method could also be applied to verify other types of tri-subject kin relations, i.e., Father/Son-Mother (FS-M), Mother/Son-Father (MS-F), Father/Daughter-Mother (FD-M), Mother/Daughter-Father (MD-F). For example, the task of FS-M is to verify whether a valid kin relation could be established between a mother and a father-son pair, given their face images.

For this series of experiments, we adopt the same 5 cross-validation evaluation protocol introduced in Section~\ref{sec_TSK} for each type of verification, with the only exception that images in each fold are partitioned according to the type of kinship of interest. We use SIFT features for face representation and follows the same parameter settings as previous experiments.

Table~\ref{tab:compare results on Different} gives results. It can be seen that the performance obtained here for different methods generally decreases by about 2-3\% compared to that in the child-parents verification (c.f., Table~\ref{tab:compare results on LFKW for tri-objects}). One possible explanation could be this: for a mixed one-vs-two relation, taking the FS-M relation for example, one has to decompose the triples of $(x_f, x_c, x_m)$ into two pairs of $(x_f, x_m)$  and $(x_c, x_m)$, and learn the pairwise similarity respectively. But the appearance similarity between a father and a mother is more difficult to learn, compared to that between a child and a mother. However, even under such a scenario, it can be seen that our proposed method (SBM-block-FS) obtains the best verification performance.

One interesting question naturally arises here is whether the appearance between a father and a mother is really similar to each other? Possibly not, because a father and a mother have different gender and have no blood relationship. But a positive father-mother pair is actually spouses who have lived together under the same living environment for a period of time, which, according to some research \cite{domingue2014genetic}, could make their appearance look more similar to each other than to others. While the size of our database is still not big enough to support this, it deserves more attention in our future research.

\subsection{The Bi-Subject Kinship Verification}
In the final series of experiments, we briefly evaluate the performance of the proposed method on the task of the bi-subject kinship verification. Particularly, we do this on two largest datasets for bi-subject verification: KinFaceW-I \cite{LuPAMI14} and KinFaceW-II \cite{LuPAMI14}. The KinFaceW-I database consists of 156 FS (Father-Son), 134 FD (Father-Daughter), 116 MS (Mother-Son) and 127 MD (Mother-Daughter) pairs, while the KinFaceW-II contains 250 pairs of these bi-subject kin relations each. The major difference between KinFaceW-I and KinFaceW-II lies in that each pair of faces in KinFaceW-I comes from the same photo while from different photos in KinFaceW-II. Hence the latter one is easier than the former.

We follow the evaluation protocol as proposed in \cite{lu2015fg}. Table \ref{tab:compare results on competition I} and Table \ref{tab:compare results on competition II} give the baseline and other latest state-of-the-art results, where the performance of the methods are directly cited from the corresponding paper. It can be seen that our original method (i.e., ``NUAA") obtains rank 3 with a simple bilinear model and without using any other features except SIFT (while both the top two methods combine several kinds of features).

It can be conjectured that combining multiple feature information could be beneficial to the performance. Hence in the next round of experiments we add two other features (i.e., the C-SVDD features~\cite{wang2013centering} and TPLBP \cite{wolf2008descriptor}) and fuse them at the decision level. This multiple feature version is denoted as ``M-NUAA" in Table \ref{tab:compare results on competition I} and Table \ref{tab:compare results on competition II}. One can see that the improved ``M-NUAA" method achieves results better than or comparable to the state of the art methods on both bi-kinship datasets, in terms of average performance (last column in both tables). Note that our algorithm is the first one designed for handle the one-vs-two tri-kinship problem and others do not, which is actually the main advantage of the proposed method: it can be thought of as a framework which can encompass any algorithm of bi-subject kinship verification for tri-subject kinship verification, while effectively incorporating useful prior knowledge.

\begin{table} [!htb]
\caption{The mean accuracy(\%) under image-restricted setting on the KinFaceW-I dataset}
\begin{center}
\begin{tabular}{l|c|lllll}
\hline
Label   &FS &FD &MS &MD & avg. \\
\hline
Polito \cite{lu2015fg}   &85.30  &\textbf{85.80} &\textbf{87.50}  &86.70  &86.30  \\
LIRIS \cite{lu2015fg}  &83.04   &80.63 &82.30 &84.98  &82.74  \\
ULPGC \cite{lu2015fg} &71.25  &70.85  &58.52  &80.89  &70.01  \\
BIU \cite{lu2015fg} &86.90  &76.48  &73.89  &79.75  &79.25  \\
NUAA(proposed)\cite{lu2015fg}  &86.25 &80.64  &81.03  &83.93  &82.96  \\
%M-NUAA(proposed)   &87.83$\pm$0.0086      &84.29$\pm$0.0078   &84.91$\pm$0.0097  &86.33$\pm$0.0098  &  \\
M-NUAA(proposed)   &\textbf{87.84}     &85.47  &86.16  &\textbf{87.50}  &\textbf{86.74}   \\
\hline
SILD(LBP)\cite{lu2015fg}  &78.22 &69.40  &66.81  &70.10  &71.13  \\
SILD(HOG)\cite{lu2015fg}  &80.46 &72.39  &69.82  &77.10  &74.94  \\
\hline
\end{tabular}
\end{center}
\label{tab:compare results on competition I}
\end{table}

\begin{table} [!htb]
\caption{The mean accuracy(\%) under image-restricted setting on the KinFaceW-II dataset}
\begin{center}
\begin{tabular}{l|c|lllll}
\hline
Label   &FS &FD &MS &MD & avg. \\
\hline
Polito \cite{lu2015fg}   &84.00   &82.20  &84.80   &81.20   &83.10   \\
LIRIS \cite{lu2015fg}  &\textbf{89.40}   &83.60  &\textbf{86.20}   &85.00   &86.05  \\
ULPGC \cite{lu2015fg} &85.40   &75.80  &75.60   &81.60   &80.00  \\
BIU \cite{lu2015fg} &87.51   &80.82  &79.78   &75.63   &80.94  \\
NUAA(proposed) \cite{lu2015fg} &84.40   &81.60  &82.80   &81.60   &82.50  \\
M-NUAA(proposed)   &88.40      &\textbf{86.20}  &86.00   &\textbf{85.20}   &\textbf{86.45     }    \\
\hline
SILD(LBP)\cite{lu2015fg}  &78.20   &70.00  &71.20   &67.80   &71.80 \\
SILD(HOG) \cite{lu2015fg} &79.60   &71.60  &73.20   &69.60   &73.50 \\
\hline
\end{tabular}
\end{center}
\label{tab:compare results on competition II}
\end{table}

\section{Conclusions}
In this work, we made the first attempt to investigate the tri-subject kinship verification problem extensively. Instead of using information from a single parent, we exploit information from both parents to learn the kinship relationship between them and their child, which is arguably one of the most important relationships formed in a family. For this we proposed a novel relative symmetric bilinear model (RSBM) and a spatially voted feature selection method, both incorporate prior knowledge about the dependence structure between a child and his/her two parents. Furthermore, we collected a new kinship face database characterized by over 1,000 groups of triples, on which we show that our method achieves state of the art verification accuracy. Our experimental results also reveal that the proposed method could be used to significantly boost the performance of bi-subject kinship verification when the information about both parents is available. Additionally, we show that our method can be applied with encouraging performance on other types of tri-subject kinship verification such as Father/Son-Mother verification, and on the traditional one-vs-one kinship problem.

Future works include further improvement based on exploiting other types of prior knowledge and learning multiple complementary features to better represent the discriminative information that is useful for our task. We also plan to extend our framework to handle more general family structure.

\para{Acknowledgements} We thank the anonymous reviewers for your in-depth comments, suggestions, and corrections, which have greatly improved the manuscript. This work was supported by the National Science Foundation of China ($61373060$, $61472186$), Jiangsu Science Foundation ($BK2012793$), Qing Lan Project, Research Fund for the Doctoral Program (RFDP)($20123218110033$), the Natural Science Foundation of the Jiangsu Higher Education Institutions of China ($13KJD520002$).

\ifCLASSOPTIONcaptionsoff
  \newpage
\fi



%\bibliographystyle{IEEEtran}
%\begin{thebibliography}{1}
%\end{thebibliography}

{\small
\bibliographystyle{elsarticle-num}
\bibliography{refbibtex}
}


%

%% insert where needed to balance the two columns on the last page with
%%\newpage


%\vfill


\end{document}